%% file: paper.tex
\documentclass[lettersize,journal]{IEEEtran}  

\usepackage[table]{xcolor}
\usepackage{graphics}    
\usepackage{times}       
\usepackage{amsmath}     
\usepackage{amssymb}     
\usepackage{graphicx}
\usepackage{algorithm}
\usepackage[noend]{algpseudocode}
\usepackage{subcaption}
\usepackage{threeparttable}
\usepackage{multirow}
\usepackage[hidelinks]{hyperref}
\usepackage{cite}
\usepackage{tikz}
\usepackage{overpic}
\usepackage{tabularx}
\usepackage{makecell}
\usepackage{booktabs}
\usepackage{stmaryrd}
\usepackage{threeparttable}

\algnewcommand\algorithmicparam{\textbf{Parameter:}}
\algnewcommand\Param{\item[\algorithmicparam]}
\input{stachnisslab-latex}

\input{stachnisslab-math}

\input{table}

\input{figure}

\usepackage[font=small]{caption}

\renewcommand{\d}[1]{\mbox{\boldmath$#1$}}	
\newcommand{\di}[1]{\mbox{\boldmath \hspace*{-0.01em}\scriptsize$#1$}}

\newcommand{\titlename}[0]{Fast and Accurate Deep Loop Closing and Relocalization for Reliable LiDAR SLAM}
\newcommand{\name}[0]{LCR-Net}
\newcommand{\textg}[1]{{\color{gray}#1}}

\newcommand{\A}[1]{{#1^{\mi A}}}
\newcommand{\B}[1]{{#1^{\mi B}}}
\newcommand{\Ai}[1]{{#1^{\mii A}}}
\newcommand{\Bi}[1]{{#1^{\mii B}}}
\newcommand{\mtf}[1]{{{\mathbf{#1}}}}

\newcommand{\mtc}[1]{{{{\mathcal{#1}}}}}
\newcommand{\hmtc}[1]{{{\hat{\mathcal{#1}}}}}

\newcommand{\bfvspace}[1]{{\vspace{0.5mm}\noindent{\textbf{#1}}}}

\makeatletter
\def\maketag@@@#1{\hbox{\m@th\normalfont\normalsize#1}}
\makeatother

\title{\titlename}

\author{Chenghao Shi$^*$, Xieyuanli Chen$^*$, Junhao Xiao$^\dag$, Bin Dai, Huimin Lu$^\dag$ 
  \thanks{C. Shi, X. Chen, J. Xiao and H. Lu are with College of Intelligence Science and Technology, National University of Defense Technology, China. B. Dai is with National Innovation Institution of Defense Technology, China. }
  \thanks{$^*$~Joint first authors with equal contribution.}
  \thanks{$^\dag$~Joint Corresponding authors: \{junhao.xiao, lhmnew\}.nudt.edu.cn}
  \thanks{
  	This work was supported in part by the National Science Foundation of China under Grant U1913202, U22A2059, and 62203460, as well as Major Project of Natural Science Foundation of Hunan Province under Grant 2021JC0004.over
  }%
}

\begin{document}
\maketitle

\markboth{}%
{Shi \MakeLowercase{\textit{et al.}}: \titlename}


\begin{abstract}
Loop closing and relocalization are crucial techniques to establish reliable and robust long-term SLAM by addressing pose estimation drift and degeneration. This article begins by formulating loop closing and relocalization within a unified framework. Then, we propose a novel multi-head network \name{} to tackle both tasks effectively. It exploits novel feature extraction and pose-aware attention mechanism to precisely estimate similarities and 6-DoF poses between pairs of LiDAR scans. In the end, we integrate our \name{} into a SLAM system and achieve robust and accurate online LiDAR SLAM in outdoor driving environments. We thoroughly evaluate our \name{} through three setups derived from loop closing and relocalization, including candidate retrieval, closed-loop point cloud registration, and continuous relocalization using multiple datasets. The results demonstrate that \name{} excels in all three tasks, surpassing the state-of-the-art methods and exhibiting a remarkable generalization ability. Notably, our \name{} outperforms baseline methods without using a time-consuming robust pose estimator, rendering it suitable for online SLAM applications. To our best knowledge, the integration of \name{} yields the first LiDAR SLAM with the capability of deep loop closing and relocalization. The implementation of our methods will be made open-source.
\end{abstract}

\section{Introduction}
\label{sec:intro}
\IEEEPARstart{S}{imultaneous} localization and mapping, also known as SLAM, plays a fundamental role across domains such as autonomous driving, robotics, and computer vision. Ensuring the reliability and stability of a SLAM system is crucial for practical applications. External sensors like GPS and IMU are commonly employed to enhance SLAM in real-world scenarios, but challenges arise when they are inaccessible or unreliable. Hence, attaining enduring reliability in LiDAR-only SLAM becomes important yet challenging. Relocalization and loop closing are pivotal techniques within this context. Relocalization refers to recovering the global pose in local tracking failure, whereas loop closing involves identifying previously visited locations to correct the drift in pose estimation. 
Despite different objectives, both techniques share similar underlying concepts. They both first coarsely locate the most similar candidate in the map relative to the current scan, followed by a fine relative pose estimation. 

More specifically, current LiDAR loop closing usually first project point clouds into image-like formats to generate global descriptors and estimate 1 degrees of freedom (DoF)~\cite{chen2021rss} or 3-DoF poses~\cite{li2021iros,chen2021auro,Harithas2023arxiv}. 
\motiFig
To achieve full 6-DoF LiDAR loop closing, multiple existing methods~\cite{steder2011iros, bosse2013icra, cui2022ral} using bag-of-words (BoW)~\cite{sivic2003iccv}. Such methods extract local features to construct BoW models and utilize such models for loop closure detection and feature matching. The final poses are then obtained by registering the matched features.
Since deep neural networks have shown great advances in perception tasks, recent studies have also employed deep learning approaches~\cite{cattaneo2022tro, Arce2022ral,Vidanapathirana2022ral} for LiDAR loop closing. 
However, they still follow a similar structure as BoW, albeit using learning-based approaches instead of hand-crafted local features extraction and BoW model.
These techniques face a dilemma: obtaining a comprehensive representation of environmental features often requires a deeper encoder. However, a deeper encoder diminishes the count of local features, potentially hindering accurate localization. Additionally, improving registration performance often involves integrating more complex designs into local feature extraction, thereby substantially reducing global description efficiency. 

Therefore, despite loop closing and relocalization sharing similar underlying techniques, few works have specifically been proposed for LiDAR relocalization. When local pose tracking degenerates, some point-based SLAM~\cite{zhang2014rss} incorporates additional sensors and switches to camera~\cite{zhang2016icra} or IMU~\cite{Xu2020ral} odometry mode, 
while some surfel-based methods~\cite{behley2018rss,chen2019iros} fall back from frame-to-map to frame-to-frame pose estimation to avoid errors caused by distorted maps. To our best knowledge, no prior research has addressed both LiDAR loop closing and relocalization simultaneously.
In this article, we aim to tackle LiDAR loop closing and relocalization using a joint framework with a novel multi-head network, named \name{}, offering four main contributions as follows:

Firstly, we revisit the challenges of addressing LiDAR loop closing and relocalization separately. We identify limitations within the current paradigm and propose \textbf{a new framework for solving loop closing and relocalization simultaneously} (\secref{sec:formulation}).
Our framework leverages the shared techniques underlying these two tasks, integrating them within a coherent coarse-to-fine framework. This framework concurrently addresses both tasks, beginning with generating global descriptors for an initial coarse global candidate search. Subsequently, the framework generates dense local features to facilitate precise 6-DoF pose estimation.
Through such a framework, we circumvent dilemmas arising from the requirements of these two tasks, providing a solid foundation for improved candidate retrieval and registration performance.

Following our framework, we then introduce \textbf{a novel multi-head network \name{}}~(\secref{sec:main}). 
It employs a shared encoder backbone to downsample and encode the point cloud into three types of sparse features. These features are then processed separately in two distinct heads. One head generates lightweight global descriptors for each scan, enabling fast candidate retrieving. The other head establishes initial sparse matches and then extends them to dense ones for accurate registration.
Unlike existing methods using only sparse features, our approach exploits fast dense feature matching based on the neighborhood consistency, achieving accurate and fast pose estimation for online applications without requiring costly robust estimator or iterative pose refinements. To effectively train the multi-head network, we also introduce novel losses with a specific training strategy, which leads to state-of-the-art~(SOTA) performance compared to all existing baselines.

To enhance the performance on both tasks, we present the third contribution as the \textbf{novel keypoint detection module~(\secref{sec:SD}) comprising two novel sub-modules, 3D-RoFormer++ enabling geometric and contextual information aggregation, and VoteEncoder for reliable keypoint detection and encoding}.
Keypoint detection module offers three types of features, enabling fast global descriptor generation and dense reliable match establishing.
Initially, it rapidly samples and encodes uniform features across the point cloud for overall representation of the environment. Subsequently, to enhance the features and make them salient and discriminative for improving registration, we introduce 3D-RoFormer++ and VoteEncoder. 3D-RoFormer++ enhances the representation capability of local features with contextual and structural information by enabling the information exchange between the two point clouds. 
The other module, VoteEncoder effectively downsamples the points while identifying keypoints lying on geometrically significant regions and subsequently aggregating the features from their neighbors. The VoteEncdoer can significantly enhance registration robustness and accuracy by improving the coverage of matching points over the overlapping area of point clouds. The superior improvement brought by VoteEncoder highlights the importance of the match distribution OVER inilier ratio, offering valuable insights~(\secref{sec:exp_whyvote}) for future research on point cloud registration. 

The fourth contribution is \textbf{the first LiDAR SLAM system with the capability of deep learning-based loop closing and relocalization}~(\secref{sec:slam}). 
We build a full LiDAR SLAM system based on our proposed deep loop closing and relocalization method. 
The system effectively tackles the local pose tracking, loop closing, and relocalization in parallel.
Enormous test of our SLAM system in diverse environments and situations showcase the effectiveness and robustness after integrating our proposed \name{}. 

We extensively evaluate our approach on three setups derived from loop closing and relocalization: candidate retrieval, closed-loop point cloud registration, and continuous relocalization. The results demonstrate that i) our approach outperforms respective baselines and dominates the SOTA in all three tasks, in particular, ii) our approach achieves the best candidate retrieval performance with a simple architecture benefiting from the feature representation capability of the backbone, iii) our approach boosts the baseline by a large margin in registration tasks, even outperforms the baseline method refined by ICP~\cite{besl1992pami}, iv) the SOTA registration performance can be achieved efficiently without requiring for a costly RANSAC estimator.
We also conduct tests on multiple sequences to assess the performance of our SLAM system. The results depict that integrating our proposed \name{}, our SLAM system is capable of addressing the relocalization and loop closing challenges. In the loop closing task, our approach outperforms the most commonly employed loop closing approach, Scan Context~\cite{kim2018iros} combined with ICP.
In addition, we provide detailed ablation studies to demonstrate the effectiveness of our design.

\section{Related Work}
\label{sec:related}
While various studies have been conducted for the foundational techniques underlying loop closing and relocalization, including candidate retrieval and point cloud registration, few work can simultaneously address both tasks.
Therefore, we first introduce the underlying techniques and then explore the recent advancements in loop closing and relocalization.

\bfvspace{Candidate retrieval}, also known as place recognition or loop closure detection, compares the current sensor observation with pre-built maps to determine the approximate location of the robot within the map. LiDAR-based loop closure detection approach can be categorized into global descriptor-based methods and local descriptor-based methods.

Global descriptor-based methods, such as Scan Context~\cite{kim2018iros}, represent point clouds as overhead views and encode different segmented spaces to construct global descriptors. Wang \etal~\cite{wang2020iros} extract descriptors using LoG-Gabor threshold filtering and measure similarity using Hamming distance. These methods require additional functions to evaluate the similarity of places, which significantly reduces computational efficiency when the map size increases.
Recent works exploited advanced deep learning techniques and generated global descriptors for more robust loop closure detection. For example, PointNetVLAD~\cite{angelina2018cvpr} first extracts local features and then aggregates them into global descriptors. Different from methods that directly operate on raw point clouds, Zhou \etal~\cite{zhou2021icra} transform point clouds into NDT and combine them with Transformer~\cite{vaswani2017nips} to generate descriptors. OverlapNet~\cite{chen2021rss} introduces a deep learning-based method, which estimates the overlap and relative yaw angles of a set of point clouds for place recognition and initial pose estimation. Ma \etal~\cite{ma2022ral,ma2023tie,ma2023tii} combine OverlapNet with Transformer to propose a rotation-invariant global descriptor. 
While the global descriptor-based methods can identify loop closures, they lack the ability to accurately estimate the 6-DoF pose between the current scan and the loop candidate.

On the other hand, local feature-based methods typically extract local sparse features from point clouds~\cite{knopp2010eccv,zhong2009iccvws} and organize them using a bag-of-words model for place recognition~\cite{bosse2013icra,cui2022ral}.
These methods are capable for 6-DoF pose estimation based on local feature correspondences.
However, extracting stable and reliable local features from 3D LiDAR scan is a challenging task, which limits the performance of such methods. 
Recently, LCDNet~\cite{cattaneo2022tro} employs a deep learning-based method PVRCNN~\cite{shi2020cvpr} for robust feature extraction and then utilizes NetVLAD~\cite{arandjelovic2016tpami} for global descriptor generation. FinderNet~\cite{Harithas2023arxiv} circumvents the challenge of feature extraction from point clouds by converting them into Digital Elevation Maps~(DEMs) and leveraging a CNN network to extract local features and global descriptors. However, converting point clouds to DEM makes it challenging to estimate 6-DoF poses.
In contrast, our method operates directly on point clouds, bypassing the challenge of estimating pose on sparse correspondences while possessing both loop closure detection and accurate 6-DoF pose estimation capability.

\bfvspace{Point cloud registration} refers to the process of determining the relative spatial transformation that aligns two point clouds. Extracting accurate correspondence is the most challenging aspect. Once correspondences are established, the transformation can be solved using either a direct solver or a robust estimator~\cite{fischler1981acm}. 
The iterative closest point (ICP) algorithm~\cite{besl1992pami} and its various variants~\cite{zhang1994ijcv, segal2009rss} are known and applied methods. These methods establish correspondences iteratively using nearest neighbor search or other heuristics.
However, the common drawback among ICP-like methods is that they heavily rely on good initial estimates for the transformation.

To release the requirement of initial estimates, other methods opt to establish correspondences on local features~\cite{rusu2009icra,zhong2009iccvws} to achieve global registration. Due to the powerful feature representation capabilities exhibited by deep learning, massive learning-based methods for feature extraction have been proposed.
Deng \etal~\cite{deng2018cvpr,deng2018eccv} propose PPFNet and PPF-FoldNet, which combine point pair features (PPF) with PointNet~\cite{qi2017cvpr} to generate local patch representations for matching.
In contrast to PPFNet and PPF-FoldNet, which establish correspondences on uniformly sampled points, keypoint-based techniques sample points based on pre-defined~\cite{zhong2009iccvws} or learned saliency~\cite{yew2018eccv,li2019iccv,bai2020cvpr,shi2021ral} to achieve better repeatability. 
Due to the inherent errors introduced by individual matches, more matches usually lead to higher registration accuracy. However, the aforementioned methods establish matches on sparse keypoints generated through uniform sampling~\cite{deng2018cvpr,deng2018eccv} or keypoint detection~\cite{yew2018eccv,li2019iccv, bai2020cvpr,shi2021ral}, which limits their registration accuracy. 
Recently, some studies~\cite{yu2021nips,qin2022cvpr,shi2023tits} employ a coarse-to-fine mechanism that initially seeks correspondences on sparse keypoints and then extends them to dense ones, showing potential in registration. To enhance the reliability of sparse keypoint correspondences, CoFiNet~\cite{yu2021nips} exploits transformer for contextual information aggregation and GeoTransformer~\cite{qin2022cvpr} introduces geometric transformer to incorporate relative geometric information. RDMNet~\cite{shi2023tits} introduces 3D-RoFormer for fast and lightweight relative geometric information encoding and the voting scheme for keypoint detection.
HRegNet~\cite{lu2021iccv} extracts multi-level features and refines the transformation hierarchically.  
Despite the rapid advancements in global registration methods, these studies have remained disconnected from the challenges of loop closing and relocalization, lacking the ability of similarity evaluation.

\bfvspace{Loop Closing and Relocalization} both need to first find a coarse location and then estimate the fine 6-DoF pose. Though the individual techniques are widely explored, as discussed in the previous review, few work can address both tasks simultaneously.
OverlapNet~\cite{chen2021auro} and FinderNet~\cite{Harithas2023arxiv} are capable of estimating 1-DoF or 3-DoF poses while implementing loop closure detection. These methods have also shown success in achieving 6-DoF loop correction when combined with other local registration techniques. However, in more challenging scenarios involving large pose drift or relocalization, non-6-DoF pose estimation is insufficient. BoW3D~\cite{cui2022ral} and LCDNet~\cite{cattaneo2022tro} achieve loop closure detection and 6-DoF pose estimation by extracting local features. Nevertheless, the accuracy of sparse local feature-based registration is constrained, requiring additional refinement through local registration techniques such as ICP. Therefore, these methods are evidently unsuitable for relocalization tasks where local registration have already failed and rapid online pose recovery are required.

To address system degeneration, most approaches integrate additional sensors such as camera~\cite{zhang2016icra}, IMU~\cite{Xu2020ral}, or ultra-wideband~(UWB)~\cite{Zhou2021tvt}, and switch between different tracking modalities.
However, to the best of our knowledge, no piror LiDAR-only method has been proposed to achieve relocalization handling system degeneracy.
This can be attributed to the challenge in achieving accurate LiDAR-based global registration when local pose tracking has already failed.

In the field of visual SLAM, however, due to the success in visual feature extraction, loop closing and relocalization have been well-explored.
ORB-SLAM~\cite{Campos2020tro} extracts ORB features to form a BoW model for loop closing. When local tracking fails, ORB-SLAM switches from frame-to-frame alignment to frame-to-map alignment to find more potential landmark matches for relocalization. OA-SLAM~\cite{Zins2022Oismar} achieves more robust relocalization by relocalizing with reconstructed objects instead of local landmarks.
Our approach, however, seeks to generate global descriptors to achieve rapid similarity evaluation for candidate detection in loop closing and relocalization. Additionally, a robust and accurate enough global registration method is employed to achieve 6-DoF pose estimation.

\frameworkFig
\pipelineFig
\section{Problem Definition}
\label{sec:formulation}
We aim to address the challenges of loop closing and relocalization for LiDAR-based SLAM in outdoor driving environments. 
The underlying techniques of relocalization and loop closing are similar: both tasks involve the identification of the most similar candidate scan from the existing map and subsequently determining the 6-DoF pose.
This commonality provides a foundation for addressing both tasks within a unified framework. 
However, the technical focus of the two tasks is quite different. Firstly, in the case of loop closing, the main challenge lies in rapidly and accurately identifying loop closures within a large database. In most cases, loop closures are identified when there is a substantial overlap between the current and candidate scans, simplifying the subsequent registration process once the loop closure has been correctly identified.
Conversely, selecting the candidate scan for relocalization is relatively straightforward. In many autonomous driving situations, simply opting for the most recent scans can be sufficient.
Instead, the primary challenge of relocalization lies in achieving precise and rapid global registration, as local pose tracking even with the aid of prior information has failed in such situations. This typically occurs in challenging scenarios that involve low overlap, extensive occlusions, or degraded scene features, posing significant challenges for point cloud registration.
Secondly, loop closing can be executed at a relatively low frequency as a few correctly closed loops are sufficient to eliminate accumulated error. However, relocalization needs to be fast as it directly affects online localization.
A longer relocalization process results in reduced overlap between the current scan and the map, decreasing the success rate of relocalization.
While numerous works have focused on loop closing, the demanding requirements of robustness, accuracy, and speed in registration could be the primary reason for the limited attention to relocalization. To address this issue, we aim to initially study the framework to support the requirements of both tasks.

A commonly employed framework for simultaneously similarity evaluation and registration is shown in \figref{fig:frame1}.
For an incoming LiDAR scan \mbox{$\mathcal{P}=\{\d p_i \in \mathbb{R}^3\}^{N}_{i=1}$}, it initially downsamples and encodes the point cloud into local features $[\hat{\mathcal{P}}|{\hat{\mathbf{F}}}]=f_{\text{Encoder}}(\mathcal{P})$ and then generate a global descriptor $\m{V}=f_{\text{Encoder}2}(\hat{\mathcal{P}},{\hat{\mathbf{F}}})$ based on these local features. The global descriptor $\m{V}$ is exploited to exhibit similarity for candidate frame retrieving and the local features $[\hat{\mathcal{P}}|{\hat{\mathbf{F}}}]$ are matched for pose estimation. 
Such a framework encounters a dilemma: A deeper encoder is often required to obtain reliable global feature representations. Nevertheless, a deeper encoder tends to yield a reduced number of local features, potentially undermine registration performance. Conversely, enhancing registration performance often entails the incorporation of more complex designs into the encoder, consequently reducing global description efficiency. This is a crucial consideration in candidate retrieval tasks.
Based on this insight, we propose the framework shown in \figref{fig:frame2}. 
We leave the workflow of global descriptor generation untouched but incorporate a decoder with skip connections for denser local feature generation ${{\mathbf{F}}}=f_{\text{Decoder}}(\hat{\mathcal{P}}|{\hat{\mathbf{F}}})$.
Though simple, this resolves the conflict between the requirements of the two tasks.
The incorporation of dense local features has the potential to enhance registration performance by establishing more correct matches, while ensuring the efficient generation of global descriptors.
However, maintaining the match quality in an increased search space can be difficult and time-consuming.
We thereby have implemented a sparse-to-dense matching approach to improve the matching process for reliable and fast registration.
By exploiting different types of features and a multi-head network, we concurrently address the disparities between loop closing and relocalization tasks while leveraging their shared characteristics to unify them within a single framework.
More detailed description of the proposed network following our framework are introduced in the next section.

\section{\name{}: Loop Closing and Relocalization Network}
\label{sec:main}
To realize our proposed framework, we design a novel multi-head network, named \name{}. As shown in~\figref{fig:model}, it consists of a keypoint detection module~(\secref{sec:SD}) to extract keypoints from the raw point cloud, a global description head~(\secref{sec:global_desc}) for global descriptor generation and a dense point matching head~(\secref{sec:dense-matching}) for local feature generation and matching. The devised loss function and the training strategy of our approach are detailed in~\secref{sec:loss} and \secref{sec:train} respectively.

\subsection{Keypoint detection module}
\label{sec:SD}
The keypoint detection module aims to downsample the point cloud into sparse keypoints for further processing in two heads. In this work, we utilize KPEncoder~\cite{thomas2019iccv} as the starting point for extracting the features. 
KPEncoder comprises a series of downsampling and kernal point-based convolution~(KPConv) blocks, enabling hierarchical encoding of the point cloud into the uniformly distributed keypoints with descriptors $[\hat{\mathcal{P}}|{\hat{\mathbf{F}}}]$.
These features provide sufficient information about the overall structure of the point cloud and are well-suited for input into the global description head. However, the uniformly sampled keypoints can not satisfy the demand for accurate registration due to their limited repeatability and saliency. They suffer from a lack of information exchange between two scans. To address these limitations and enhance feature matching, we introduce the \textit{3D-RoFormer++} to reason about contextual information in both point clouds. 
In addition, we propose a new \textit{VoteEncoder} that shifts the keypoints to nearby significant regions based on enhanced features and generates the final keypoints by predicting the center points.
We provide detailed descriptions of each component below.

\bfvspace{3D-RoFormer++.}
In our previous work, the 3D-RoFormer~\cite{shi2023tits} is introduced for lightweight relative pose-aware contextual aggregation.
In this article, we have brought 3D-RoFormer to maturity and present the 3D-RoFormer++ by providing valuable translational invariance and enhanced feature representation performance.  
The 3D-RoFormer is built upon the vanilla transformer~\cite{vaswani2017nips}. 
For a point $\d p_i^{\mi Q}$ with its feature $\d h_i^{\mi Q}$ in the query point cloud $\m Q$ and all the points in the source point cloud $\m S$, the transformer computes the query~$\d q_i$, key~$\d k_j$, and value~$\d v_j$ feature maps with linear projections:
\begin{align}
\label{eq:attention}
\d q_i &= \m W_1 \, \d f_i^{\mi Q}+\d b_1,
\nonumber\\ 
\d k_j &= \m W_2 \, \d f_j^{\mi S}+\d b_2,
\\
\d v_j &= \m W_3 \, \d f_j^{\mi S}+\d b_3. \nonumber
\end{align} 
If $\m Q,\m S$ represent the same point cloud, \eqref{eq:attention} generates the feature maps for self-attention operation, otherwise cross-attention.
In addition to the contextual features, 3D-RoFormer encodes the position $\hat{\d p}_i\in \mathbb{R}^3$ into the rotary embedding \mbox{$\d \Theta_i = [\theta_1, \theta_2, \cdots, \theta_{d/2}] \in \mathbb{R}^{ \frac{{d}}{2}}$}:
\begin{align}
\d \Theta_i&=f_{\text{rot}}(\hat{\d p}_i))\\ 
&=2\pi\cdot\text{sigmoid}(\text{MLP}(\hat{\d p}_i)).
\label{eq:mapping}
\end{align}
By treating each element in $\d \Theta_i$ as a rotation in a 2D plane, it can be converted to a rotation matrix formulation \mbox{$\m R_{ \d \scriptsize\Theta_i}\in\mathbb{R}^{{d}\times {d}}$}:
\begin{small}
	\begin{align}
	\label{eq:rope_r}
	\m R_{ \d \scriptsize\Theta_i}
	&=
	\begin{bmatrix} &\cos \theta_1\!&-\sin \theta_1\!&\cdots\!&0\!&0
	\\ &\sin \theta_1\!&\cos \theta_1\!&\cdots\!&0\!&0
	\\&\vdots\!&\vdots\!&\ddots\!&\vdots\!&\vdots
	\\&0\!&0\!&\cdots\!&\cos \theta_{\frac{d}{2}}\!&-\sin \theta_{\frac{d}{2}}
	\\&0\!&0\!&\cdots\!&\sin \theta_{\frac{d}{2}}\!&\cos \theta_{\frac{d}{2}}
	\end{bmatrix}.
	\end{align}
\end{small}
Applying $\m R_{ \d \scriptsize\Theta_i}$ and $\m R_{ \d \scriptsize\Theta_j}$ to query~$\d q_i$ and key~$\d k_j$ respectively in self-attention operation, the rotary self-attention in 3D-RoFormer can be written as:
\begin{align}
\label{eq:roformer}
\alpha_{ij}''&=\textrm{softmax}_j((\m R_{ \d \scriptsize\Theta_i} \d q_i)\trans\m R_{ \d \scriptsize\Theta_j}\d k_j),\\
\tilde{\d f}_i
&=\sum_{j=1}^{|\hat{\mathcal{P}}|}\alpha_{ij}''\d v_j. \label{eq:roformer3}
\end{align}
The \eqref{eq:roformer} can be further written as:
\begin{align}
\label{eq:roformer2}
\alpha_{ij}''
&=\textrm{softmax}_j(\d q_i\trans\m R_{ \d \scriptsize\Theta_i}\trans\m R_{ \d \scriptsize\Theta_j}\d k_j),\nonumber\\
&=\textrm{softmax}_j(\d q_i\trans\m R_{ \d \scriptsize\Theta_j-\d \scriptsize\Theta_i}\d k_j).
\end{align}

The important advantage of 3D-RoFormer is that it explicitly encodes the relative geometric information neatly without requiring extra-large storage memory for relative position embedding.
As in \eqref{eq:roformer2}, relative ``rotation" $\d \Theta_j-\d \Theta_i$ is naturally incorporated into the calculation and then fused with the output feature $\tilde{\d f}_i$ in \eqref{eq:roformer3}. 
Furthermore, if the mapping function $f_{\text{rot}}$ is linear, we can further derive:
\begin{align}
\label{eq:linearity}
\d \Theta_j-\d \Theta_i = f_\text{rot}(\hat{\d {p}}_j-\hat{\d p}_i).
\end{align}
This leads to a very important property for {keypoint detection} which is translation-invariance.
However, designing a $f_{\text{rot}}$ that provides good rotary feature representation while maintaining linearity is challenging.
To ensure the ability of rotary representation, the rotary embedding in the original 3D-RoFormer, as shown in~\eqref{eq:mapping}, sacrifices linearity for rotary representation, leading to reduced generalization performance.

Based on this insight, we improve our 3D-RoFormer by adopting a learning-based linear mapping function:
\begin{align}
\label{eq:linear_mapping}
\d \Theta_i=\text{Linear}(\hat{\d p}_i),
\end{align}
with a boundary penalty loss~(\secref{sec:loss}) as an auxiliary loss to supervise the network actively learning effective rotary representations.
With this modification, our 3D-RoFormer++ significantly enhances the final output features $\tilde{\mathbf{F}}^{\mi A}$ and $\tilde{\mathbf{F}}^{\mi B}$ for point matching by interleaving the rotary self-attention and cross-attention for $l$ times.

The enhanced features $\tilde{\mathbf{F}}$ possess geometric and contextual information between two point clouds, which is then extended to dense features for further processing in dense point matching head, as detailed in~\secref{sec:dense-matching}.
Nevertheless, the uniform sampling nature makes these features less salient and discriminative. We therefore propose the VoteEncoder.

\voteencoderFig

\bfvspace{VoteEnoder}. To steer the evenly sampled features $[\hat{\mathcal{P}}|{\tilde{\mathbf{F}}}]$ towards nearby salient areas and obtain more meaningful features conducive to registration tasks, we introduce the VoteEncoder for additional feature encoding.

We use a voting module~\cite{Qi2019iccv,shi2023tits} to estimate the geometric offset from the uniformly sampled keypoints to the proposal keypoints $\mtc{S}$, i.e., \mbox{$\Delta{\mathbf{P}}=\text{Vote}(\tilde{\mathbf{F}})$}, $\mtc{S}=\hat{\mathcal{P}}+\Delta\mathbf{P}$. 
The voting module comprises a collection of Multi-Layer Perceptrons (MLPs). Despite its simplicity, this module produces meaningful offsets (see~\figref{fig:voteencoder}), utilizing the features from our 3D-RoFormer++. These generated proposals subsequently forecast multiple central points, serving as the final keypoints. The process of predicting centroids is straightforward yet efficient, without the need for additional sampling strategies. It clusters all the proposals into various patches and predicts the centers, detailed in~\algref{alg:centroid}.
To aggregate the descriptors ${\mathbf{H}}$ for each center point $\hat{\mtc{S}}$, we employ a KPConv module that performs kernel-based convolution after finding nearest neighbors of $\hat{\mtc{S}}$ in $[\hat{\mathcal{P}}|{\tilde{\mathbf{F}}}]$.
We use a larger search range than that used in \algref{alg:centroid} to incorporate more related context near the keypoints.

Unlike the object detection~\cite{Qi2019iccv,Zhang2022cvpr}, where the object center can serve as a well-defined reference for supervising point shifts, our case lacks a readily available ground truth center for the significant areas, primarily due to the challenge in precisely defining the significance. Therefore, we train the matched keypoints to move closer to each other instead, which indirectly accomplishes our objective.
In practice, the offset $\Delta\mathbf{P}$ is limited to a certain range to maintain an even distribution of keypoints throughout the point cloud. This prevents the keypoints from being only concentrated in significant areas while also avoiding potential degeneracy.

\begin{algorithm}[t]
		\caption{Centroid prediction}
	\begin{algorithmic}[1]
		\Require{proposal set $\mathcal{S}$, nearest neighbour search range $d$}
		\Ensure{center point set $\hat{\mathcal{S}}$}
		\For {all $\d s_i \in \mathcal{S}$}
		\If {$\Call{Islabeled}{\d s_i}$ is False}
		\State $\mathcal{N}_i \gets \Call{NearestNeighbour}{\d s_i,\mathcal{S},d}$
		\State $\hat{\d s}_i \gets \Call{Mean}{\mathcal{N}_i}$ \Comment{Get centroid point}
		\State $\hat{\mathcal{S}}$.\Call{Append}{$\hat{\d s}_i$}
		\For {all $\d s_j \in \mathcal{N}_i$}
		\State $\Call{Islabeled}{\d s_j} \gets \text{True}$
		\EndFor

		\EndIf
		\EndFor
		\State \textbf{return} $\hat{\mathcal{S}}$
\vspace{-0.1cm}
	\end{algorithmic}\label{alg:centroid}
\end{algorithm}

In sum, our keypoint detection module offers various options for the sparse features, including uniformly sampled features $\hat{\mathbf{F}}$ from KPEncoder~\cite{thomas2019iccv}, enhanced features $\tilde{\mathbf{F}}$ from 3D-RoFormer++ and voted features $\mathbf{H}$ from VoteEncoder. 
Uniformly sampled features $\hat{\mathbf{F}}$ are distributed evenly throughout the entire point cloud, enabling fast extraction and the capacity to represent the entire point cloud comprehensively. These features are utilized for the global description head, as detailed in \secref{sec:global_desc}. 
On the other hand, enhanced features $\tilde{\mathbf{F}}$, built upon the uniformly sampled features $\hat{\mathbf{F}}$, incorporate geometric information and the correlation between two scans. These features will be decoded into dense features, propagating the aforementioned advantages to them.
Finally, voted features $\mathbf{H}$ exhibit sensitivity and expressive capability for local salient regions while maintaining good coverage. These features will be employed for initial sparse matching, which will then be extended to dense matching for accurate registration.
Both enhanced features and voted features will be employed for the dense point matching head detailed in \secref{sec:dense-matching}.

\subsection{Global Description Head}
\label{sec:global_desc}
The objective of the global description head is to condense sparse features into a single global feature for fast candidate retrieval. 
We adopt the features derived from the KPEncoder as the input to our global description head. 

When creating the global description head, we choose a widely applied simple method NetVLAD~\cite{arandjelovic2016tpami} to compress the features, and a context gating module~\cite{cattaneo2022tro} to enhance these features for retrieval.
NetVLAD uses k-means clustering and defines $K$ learnable cluster centers $\{\d{c}_1,\cdots,\d{c}_K\}, \d{c}_k\in\mathbb{R}^{\hat{d}}$, along with learnable weights $\d{w}_k$ and offsets $b_k$. By weighting each feature to each cluster center:
\begin{eqnarray}
	a_k(\hat{\mathbf{F}}_i)=
	\frac{e^{\d{w}_k\trans\hat{\mathbf{F}}_i+b_k}}{\sum_{k'=1}^{K}e^{\d{w}_{k'}\trans\hat{\mathbf{F}}_i+b_{k'}}},
\end{eqnarray}
the new descriptors for $K$ cluster centers are obtained:
\begin{eqnarray}
{\mathbf{FR}}=[\mathbf{FR}_1,\cdots,\mathbf{FR}_K]\in\mathbb{R}^{K\times\hat{d}}, \\
\mathbf{FR}_K=\sum_{i=1}^{|\hat{\mathcal{P}}|}a_k(\hat{\mathbf{F}}_i)(\hat{\mathbf{F}}_i-\d{c}_k).
\end{eqnarray}

Finally, a simple MLP compresses $\mathbf{FR}$ into a single descriptor $\m X\in\mathbb{R}^{G}$.
Based on NetVLAD, the context gating module re-evaluates the weights of each channel of feature $\m X$ based on the self-attention mechanism and further enhances it to obtain the final global descriptor $\m{V}\in\mathbb{R}^{G}$:
\begin{eqnarray}
	\m{V}=\text{CG}(\m X)=\sigma(\m W\m X+\d b)\otimes\m X,
\end{eqnarray}
where $\sigma$ is the sigmoid activation function, $\otimes$ is element-wise multiplication, and $\m W$ and $\d b$ are learnable weights and offsets.

The design of the global description head is straightforward yet remarkably effective, surpassing all baseline methods in our experiments. 
Its simplicity is also particularly important because LiDAR SLAM requires quick and accurate retrieval for real-time canditate retrieval, which narrows down the computational scope for subsequent fine 6-DoF pose estimation.

\subsection{Dense Point Matching Head}
\label{sec:dense-matching}
Once identifying the candidates, we leverage the dense point matching head to establish correspondences and subsequently recover precise 6-DoF pose estimation. 
The features obtained from our keypoint detection module $[\hmtc{S}|\mtf{H}]$ are sufficient for ensuring dependable point cloud registration. However, there are two factors that impact the accuracy of the final 6-DoF pose estimation. Firstly, despite VoteEncoder improving keypoint locations, there might still be noticeable distances between matched sparse features. These gaps can lead to errors that restrict the overall accuracy. Secondly, due to the sparse characteristics of these features, there might not be adequate feature matches to fully rectify errors arising from mismatches.
Considering these limitations, we employ a two-step matching approach~\cite{yu2021nips,qin2022cvpr}. Initially, we identify sparse yet dependable keypoint matches, and then we extend these point-to-point matches to patch-to-patch matches. By utilizing neighbor consistency, we enhance these patch matches into dense point matches, ensuring more precise and reliable registration.

\bfvspace{Sparse keypoint matching.}
We firstly conduct sparse keypoint matching between $[\hmtc{S}\A{}|\mtf{H}\A{}]$ and $[\hmtc{S}\B{}|\mtf{H}\B{}]$.
We compute a matching score matrix ${\mathbf{C}}\in\mathbb{R}^{|\hat{\mathcal{S}}^{\mii A}|\times|\hat{\mathcal{S}}^{\mii B}|}$ between ${\mathbf{H}}^{\mi A}$ and ${\mathbf{H}}^{\mi B}$:
\begin{align}
\label{eq:node_match_score}
\mathbf{C} = \A{\mathbf{H}}(\B{\mathbf{H}})\trans/\sqrt{{d_c}},
\end{align} 
where $d_c$ refers to the feature dimension of $\mathbf{H}$.
To handle non-matched points, we append a ``dustbin" row and column for $\mathbf{C}$ filled with a learnable parameter $\alpha\in\mathbb{R}$. The Sinkhorn algorithm~\cite{Sinkhorn1964ams} is then used to solve the soft assignment matrix. It iteratively performs normalization along rows and columns. At the $t$ iteration, the score matrix is updated by:
\begin{align}
\label{eq:c_OT}
{}^{(t)}{\mathbf{C}}'_{ij} &= {}^{(t)}{\mathbf{C}}_{ij} - \log\sum_{j}e^{{}^{(t)}{\mathbf{C}}_{ij}},\\
{}^{(t+1)}{\mathbf{C}}_{ij} &= {}^{(t)}{\mathbf{C}}'_{ij} - \log\sum_{i}e^{{}^{(t)}{\mathbf{C}}'_{ij}}.
\end{align} 

After $T$ iterations, we use the solution as the soft assignment matrix: $\hat{\mathbf{C}} = {}^{(T)}{\mathbf{C}}$.
We choose the largest $N_c$ entries as the keypoint correspondences:
\begin{align}
\label{eq:superpoint_corres}
{\mtc{C}} = \{(\A{\hat{\d s}_{x_i}},\B{\hat{\d s}_{y_i}})|(x_i,y_i)\in\text{Top-k}_{x,y}(\hat{\mathbf{C}})\}.
\end{align} 

\bfvspace{Patch grouping.}
To achieve dense point matches from sparse keypoint matches, we expand correspondences between keypoints to encompass overlaps between their respective neighborhood patches and subsequently leverage these patches to identify more point matches. 

For each keypoint $\hat{\d s}_i$, we construct a local patch ${\mathcal{G}_i}$ using a point-to-node strategy~\cite{li2019iccv}, where each point is assigned to its nearest keypoint.
Based on the grouped point patch, we can now extend each keypoint match $(\A{\hat{\d s}_{x_i}},\B{\hat{\d s}_{y_i}})$ to its corresponding patch match $(\A{\mtc G_{x_i}}, \B{\mtc G_{y_i}})$.

\bfvspace{Dense point matching.}
We then generate more point matches from the sparse patch matches. 
We leverage the KPDecoder~\cite{thomas2019iccv} to recover point-level descriptors $\mtf F$ from enhanced keypoint features $[\hat{\mathcal{P}}|{{\tilde{\mtf F}}}]$. 
For each keypoint correspondence $(\A{\hat{\d s}_{x_i}},\B{\hat{\d s}_{y_i}})$, we compute a match score matrix ${\mathbf{O}_i}\in\mathbb{R}^{|{\mathcal{G}}_{x_i}^{\mi A}|\times|{\mathcal{G}}_{y_i}^{\mi B}|}$ of their corresponding patches $\A{\mtc G_{x_i}}$ and $\B{\mtc G_{y_i}}$:
\begin{align}
\label{eq:OT}
\mathbf{O}_i = \A{\mtf{F}_{x_i}}(\B{\mtf{F}_{y_i}})\trans/\sqrt{{d_f}},
\end{align}
where $d_f$ refers to the feature dimension of $\mtf F$.
Same with our sparse keypoint matching module, we append a learnable ``dustbin" row and column for $\mathbf{O}_i$ to handle non-matched points and use the sinkhorn algorithm to solve the soft assignment matrix $\mtf{Z}_i\in\mathbb{R}^{(|{\mathcal{G}}_{x_i}^{\mi A}|+1)\times(|{\mathcal{G}}_{y_i}^{\mi B}|+1 )}$.
Unlike works~\cite{qin2022cvpr,yu2021nips} that drops the dustbin and recovers the assignment by comparing the soft assignment score with a hand-tuned threshold, we directly find max entry both row-wise and column-wise on $\mtf{Z}_i$ which is then recovered to assignment $\mtc{M}_i$:

\begin{small}
\vspace{-0.2cm}
	\begin{align}
	\label{eq:dense_match}
	\mtc{M}_i
	=&\{(\A{\mtc G_{x_i}}(m),\B{\mtc G_{y_i}}(n)|(m,n)\in \text{toprow}_{m,n}(\mtf{Z}_{1:M_i,1:(N_i+1)}^i)\}\cup\nonumber\\
	&\{(\A{\mtc G_{x_i}}(m),\B{\mtc G_{y_i}}(n)|(m,n)\in \text{topcolumn}_{m,n}(\mtf{Z}_{1:(M_i+1),1:N_i}^i)\}.
	\end{align} 
\end{small}
A point is either assigned to points in the matched patch or to the dustbin.
By this, we do not need manual tuning but require a discriminative assignment matrix, which can be obtained by using our proposed loss function as detailed in \secref{sec:loss}.
Note that a point is not strictly assigned to a single point in our approach, as the strict one-to-one point correspondences do not hold in practice due to the sparsity nature of the LiDAR scans. 
Instead, we trust and keep the assignment results from both sides, i.e., matches from query to source and vice versa.
This results in extensively more point matches while maintaining a high inlier ratio, which benefits the transformation estimation.
The final correspondences are the combination of points matches from all patches:
\begin{align} 
	\mtc{M}=\bigcup_{i=1}^{N_c}\mtc{M}_i.
\end{align}

\bfvspace{Local-to-global registration.}
We use local-to-global registration~(LGR) proposed in~\cite{qin2022cvpr} for fast pose estimation. It is a hypothesize-and-verify approach specifically proposed for matching methods following a sparse-to-dense manner.
For each matched patch, LGR solves a transformation $\{\m R_i, \d t_i\}$ based on its dense point matches using weighted SVD~\cite{besl1992pami}:
\begin{align}
\label{eq:rt}
\m R_i, \d t_i=\min_{\m R,\d t}\sum\limits_{\tiny{(\Ai{\d p_{x_j}},\Bi{\d p_{y_j}})\in\mtc{M}_i}}{\omega_j^i\|\m R\cdot\A{\d p_{x_j}}+\d t - \B{\d p_{y_j}}\|_2^2},
\end{align}
where the soft assignment value in $\mtf{Z}^i$ serves as the weight $\omega_j^i$.
After obtaining the transformations for all matched patches, LGR selects the transformation that has the most inliers among all dense point matches:
\vspace{-0.3cm}

\begin{small}
\begin{align}
\m R, \d t=\max_{\m R_i,\d t_i}\sum\limits_{\tiny{(\Ai{\d p_{x_j}},\Bi{\d p_{y_j}})\in\mtc{M}}}\llbracket{\|\m R_i\cdot\A{\d p_{x_j}}+\d t_i - \B{\d p_{y_j}}\|_2^2<\tau_a}\rrbracket,
\end{align}
\end{small}
where $\llbracket\bullet\rrbracket$ is an indicator function for which the statement is true.
Finally, it solves the final transformation $\m R, \d t$ by solving \eqref{eq:rt} on surviving inliers for $N_r$ times.

LGR significantly reduces the number of iterations compared to RANSAC~\cite{fischler1981acm}, achieving a substantial speed advantage with about 30 times faster in our experiments. However, the performance of LGR, particularly its robustness, can be heavily influenced by the quality of sparse patch matching. We significantly improve the matching quality of sparse patches through the powerful feature aggregation module 3D-RoFormer++ and the feature detection module VoteEncoder, achieving performance comparable to or surpassing RANSAC's accuracy and robustness.

\subsection{Loss function}
\label{sec:loss}
To effectively guide our network in accomplishing various tasks, we construct our loss function with five components: the keypoint detection loss $L_\text{s}$, the boundary penalty loss for keypoint detection module, the triplet loss $L_\text{t}$ for global description head, and the sparse match loss $L_\text{c}$ and the dense match loss $L_\text{f}$ for dense point matching head.

\bfvspace{Keypoint detection loss.} The keypoint detection loss consists of two parts $L_{s}=L_{s1}+L_{s2}$. The first part $L_{s1}$ is designed to guide the corresponding keypoints from two point clouds close to each other lying within the significant region:
\begin{equation}
	\label{eq:ls1}
	L_{\text{s}1}=\sum_{i=1}^{|\mtc{S}^{\mii A}|}\min_{\di s^{\mii B}_j\in\mtc{S}^{\mii B}}\|\d s^{\mii A}_i-\d s_j^{\mii B}\|^2_2+\sum_{i=1}^{|\mtc{S}^{\mii B}|}\min_{\di s_j^{\mii A}\in\mtc{S}^{\mii A}}\|\d s_i^{\mii B}-\d s_j^{\mii A}\|^2_2.
\end{equation}
Supervised by $L_{\text{s}1}$, we find that the keypoints tend to move to their nearest significant regions to indirectly minimize the distance between keypoint pairs.

The second part $L_{s2}$ is designed to make the keypoints close to the real measurement points. It minimizes the distance between the keypoint with its closest point:
\begin{equation}
	\label{eq:ls2}
	L_{s2}=\sum_{i=1}^{|\mtc{S}^{\mii A}|}\min_{\di p^{\mii A}_j\in\mtc{P}^{\mii A}}\|\d s^{\mii A}_i-\d p_j^{\mii A}\|^2_2+\sum_{i=1}^{|\mtc{S}^{\mii B}|}\min_{\di p_j^{\mii B}\in\mtc{P}^{\mii B}}\|\d s_i^{\mii B}-\d p_j^{\mii B}\|^2_2.
\end{equation} 

\bfvspace{Boundary penalty loss.}
To guide the 3D-RoFormer++ learning a general rotary embedding representation, we add a boundary penalty loss to force the value of rotary embedding $\d \Theta$ lies between $[-\pi,\pi]$:
\begin{align}
	\label{eq:penalty_loss}
	L_{\text{p}}^i =&\frac{1}{M_i}\sum_{m=1}^{M_i}[\text{abs}(\d \Theta)-\pi]_+.
\end{align}

\bfvspace{Triplet loss.}
We use the triplet loss to train the global description head. 
For each scan, we define the scans with an overlap greater than 30\% as positive, otherwise negative.
For each triplet, we use one query scan, $N_p$ positive scans and $N_n$ negative scans. The triplet loss is calculated as:
\begin{align}
	\label{eq:trip}
	&L_{\text{t}}(\m V_q,\{\m V_p\},\{\m V_n\})=\\ \nonumber
	&N_p(\alpha+\max_p{(d(\m V_ q,\m V_p))}-\frac{1}{N_n}\sum_{N_n}(d(\m V_q,\m V_n))).
\end{align}

\bfvspace{Sparse match loss.}
We utilize a gap loss~\cite{shi2021ral} to learn a discriminative soft assignment matrix $\mathbf{C}$ for sparse keypoint matching. 
The ground truth of assignment matrix \mbox{$\mtf{P} \in \{0,1\}^{(M{+}1)\times(N{+}1)}$} is generated based on the overlap ratio between the patches, where $M=|\A{\hat{\mtc{S}}}|$ and $N=|\B{\hat{\mtc{S}}}|$ are the keypoints number.
Two patches are matched when they share at least 10\% overlap.
A patch is assigned to the dustbin when it has no match pair.
We also generate a negative assignment matrix {$\bar{\mtf{P}} \in \{-\inf,1\}^{(M{+}1)\times(N{+}1)}$, where $1$ represents two patches are not overlapped and $\inf$ represents the value will not involve in the calculation of loss.}
Then the gap loss is calculated as:
\begin{align}
	\label{eq:node_gap}
	L_{\text{c}} &=\frac{1}{M}\sum_{m=1}^{M}\log(\sum_{n=1}^{N+1}(- r_{m}+\mtf{C}_{m,n}\bar{\mtf{P}}_{m,n}+\eta)_+ +1) \nonumber\\ &+\frac{1}{N}\sum_{n=1}^{N}\log(\sum_{m=1}^{M+1}(- c_{n}+\mtf{C}_{m,n}\bar{\mtf{P}}_{m,n}+\eta)_+ +1),
\end{align}
{where
	\begin{align}
		r_{m} = \max_m(\mtf{C}_{m,n}{\mtf{P}_{m,n}}), \quad c_{n} = \max_n(\mtf{C}_{m,n}{\mtf{P}_{m,n}}),
	\end{align}
	refer to the soft assignment value for the hardest true match of \mbox{$m$-th} keypoint in $\A{\hat{\mtc{S}}}$ and \mbox{$n$-th} keypoint in $\B{\hat{\mtc{S}}}$ respectively, and \mbox{$(\bullet)_+=\max(\bullet,0)$}.}

\bfvspace{Dense match loss.}
The dense match loss is calculated over all the matched patches. For each matched patch pair $\{\A{\mtc G_{x_i}},\B{\mtc G_{y_i}}\}$, we generate its ground truth positive correspondences matrix \mbox{$\mtf{M}^i \in \{0,1\}^{(M_i{+}1)\times(N_i{+}1)}$} and negative matrix \mbox{$\bar{\mtf{M}}^i \in \{10^{12},1\}^{(M_i{+}1)\times(N_i{+}1)}$} with a distance threshold $\tau$, where $M_i=|\A{\mtc G_{x_i}}|$, $N_i=|\B{\mtc G_{x_i}}|$. A point pair is positive when the distance is below $\tau$ and is negative when it exceeds $2\tau$.
To learn a discriminative soft assignment matrix, we also calculate a gap loss $L_{\text{f}}^i$ for patch correspondence's soft assignment matrix $\mtf{Z}^{i}$.
The final fine match loss is the average over all the matched patch pairs: \mbox{$L_\text{f}=\frac{1}{2|\mtc{M}|}\sum_{i=1}^{|\mtc{M}|}L_{\text{f}}^i$}.

\subsection{Training Strategy}
\label{sec:train}
We seek a training strategy to stimulate the potential of each head with limited computing resources.
As a result, a two-stage training strategy is employed. 
We first train the keypoint detection module and the dense point matching head for registration. Then, we exclusively train the global description head for the candidate retrieval for the following reasons:

Firstly, the input for the training of two heads differs. The training of the global description head requires at least three scans: an anchor, a positive, and a negative. Conversely, training the dense point matching head only requires two overlapped scans. Including additional input does not benefit the training of the dense point-matching head but consumes a significant amount of memory. 
Secondly, for tasks of candidate retrieval, a higher batch size typically results in better performance. By freezing the keypoint detection module and dense point matching head, we can use the pre-extracted features for input, thereby preserving substantial memory for expanding the batch size.

However, using pre-extracted features prevents the implementation of data augmentation techniques, thereby limiting performance. To address this problem, we utilize a training strategy, which we refer to as semi-online. In this approach, we utilize offline pre-extracted features for both positive and negative samples while generating features for the anchor online. This allows for applying data augmentation on the anchor. Since the anchor participates in loss calculations with all positive and negative samples, this can be the most efficient way to implement data augmentation.

In sum, our two-stage training strategy first trains the network in the registration task and then exclusively trains the global description head semi-online for the candidate retrieval task. This strategy offers several advantages. Firstly, it allows us to utilize larger batch sizes and sample quantities during training for the candidate retrieval task, leading to improved results. Secondly, the semi-online approach enables the application of data augmentation techniques. Lastly, this strategy provides great convenience in training as we only need to select the pre-trained network that performs best in registration for the second-stage training, followed by selecting the network that performs best in the candidate retrieval task. These advantages contribute to this training strategy's enhanced performance compared to fine-tuning on a pre-trained model and end-to-end training, as demonstrated in our experiments.

We implement and train our \name{} on $4$ NVIDIA RTX 3090 GPUs. 
The network is trained using the Adam optimizer~\cite{kingma2014arxiv} with an initial learning rate as $10^{-4}$, which undergoes exponential decay by $0.05$ every $4$ epochs. 
When training the dense point matching head, we use a batch size of $1$. On the other hand, when training the global description head, we set the batch size to $6$, utilizing $N_a=1$ anchor scan, $N_p=6$ positive scans, and $N_n=6$ negative scans.
Additionally, we apply the same data augmentation techniques as in~\cite{cattaneo2022tro}.

\section{Robust Loop Closing and Relocalization Based LiDAR SLAM System}
\label{sec:slam}
\slamFig
We integrate our \name{} into a SLAM system for relocalization and loop closing.
We use an Incremental Smoothing and Mapping~(iSAM2)~\cite{kaess2012ijrr} based pose-graph optimization~(PGO)~\footnote{https://github.com/gisbi-kim/SC-A-LOAM.} framework to manage and optimize the global pose graph.
The system comprises three threads that run in parallel: tracking, relocalization, and loop closing. 

\subsection{Tracking}
The tracking thread is responsible for tracking the LiDAR pose of every scan, i.e., odometry. It also decides whether the current odometry estimation is degenerated and when to insert a new keyframe. The process of odometry can be formulated as a state estimation problem:
\begin{align}
\label{eq:state_est}
\arg\min_{\d x}{f^2(\d x)},
\end{align}
where $\d x=[\m R, \d t]$ is the state vector. Given an initial guess of $\d x$, most nonlinear optimization methods solve the function by computing the Jacobian matrix of $f$ w.r.t $\d x$:
\begin{align}
\label{eq:jaco}
\m J=\delta f(\d x)/\delta \d x,
\end{align}
and iteratively adjust $\d x$ by utilizing $\m J$ until convergence.
The tracking thread can be implemented through LiDAR odometry, such as LOAM~\cite{zhang2014rss}.
For degeneracy evaluation, we adopt the criterion proposed in~\cite{zhang2016icra}, which considers the problem degenerate if the smallest eigenvalue of matrix $\m J\trans \m J$ falls below a certain threshold. 
In practice, the threshold is conservatively set to ensure that all degeneracy can be detected.
Though this leads to more frequent relocalization and increased computational load, it is necessary since even a single instance of severe degeneracy could inflict significant damage on the system.

LiDAR scans are keyframes if the robot has moved beyond a predefined threshold from the last saved keyframe. 
However, when odometry performance deteriorates, relying solely on the previous travel distance to establish keyframes becomes unreliable. To address this, we introduce additional scans as keyframes based on specific conditions as follows:
i) The first degenerated scan follows nondegenerated scans.
ii) The first nondegenerated scan follows a degenerated scan, provided the degenerated scan is not chosen as a keyframe.
iii) Every $t$-second scan within a continuous sequence of degenerated scans.
The keyframes selected according to these conditions are categorized as ``degenerated'', as poses estimated between these keyframes may include degenerated ones.
Finally, for each keyframe, we use \name{} to generate a global descriptor.
The similarity between our generated global descriptors is evaluated based on the Euclidean distance in the feature space.  
For fast retrieving, we utilize the FAISS library~\cite{Johnson2017tbd} for descriptor database management and search.

\subsection{Relocalization}
The relocalization processes every ``degenerated'' keyframe, generating more accurate pose estimation to replace the unreliable odometry output.
We select the most recent ``nondegenerated'' keyframe for each degenerated keyframe as the candidate and match it with the current keyframe using \name{}. For real-time efficiency, we utilize the rapid estimator LGR for pose estimation. Our experimental results show that employing \name{} with LGR produces comparable accuracy and robustness in registration compared to \name{} with RANSAC while offering a significantly faster processing speed of nearly $30$ times. 
To assess whether the relocalization is successful, we calculate a current pose estimation reliability score by averaging the assignment score of all the found correspondences. The relocalization succeeds if the reliability score surpasses a threshold $\rho_r$.
Once the calculated pose is included in the graph optimization, the label assigned to the current keyframe is updated to ``nondegenerated''. In most cases, this allows for recovery of the pose tracking. Otherwise, we retrieve the most similar candidate with descriptor distance below a threshold $\rho_s$ from the database and attempt relocalization.

\subsection{Loop closing}
The loop closing thread functions by searching the loop candidate and subsequently estimating pose as a new node added to the pose graph.
Specifically, we retrieve the candidate keyframe from the database that has the most similar descriptor for each new keyframe while excluding the $100$ most recent keyframes. 
If the distance is below a threshold $\rho_s$, we set the retrieved keyframe as the loop and estimate the relative 6-DoF pose using \name{} with the LGR solver. Finally, the iSAM2-based pose graph optimization is performed to achieve global consistency.


\section{Experimental Evaluation}
\label{sec:exp}

We conduct experiments to demonstrate the efficacy of our proposed \name{} in addressing loop closing and relocalization for online LiDAR SLAM. 
We derive three setups from these challenges: candidate retrieval, closed-loop point cloud registration, and continuous relocalization.
In candidate retrieval experiments~(\secref{sec:exp_lcd}), we examine the capability of \name{} in accurately retrieving the appropriate candidates from the previous map.
In closed-loop point cloud registration~(\secref{sec:exp_lc}) and continuous relocalization experiments~(\secref{sec:exp_con_regis}), we assess the ability of \name{} in successfully registering point clouds during loop situations as well as continuous scenarios with low overlap.
We also evaluate the runtime of our approach for candidate retrieving and point cloud registration~(\secref{sec:exp_runtime}).
We then evaluate our \name{} enhanced SLAM in multiple real-world scenes.
Finally, we conduct ablation studies on the network design~(\secref{sec:exp_ab}) and provide valuable insights~(\secref{sec:exp_whyvote}).

\subsection{Experimental Setup}
We evaluate \name{} and compare it with the SOTA methods on multiple publicly available datasets, including KITTI odometry~\cite{geiger2012cvpr}, KITTI-360~\cite{liao2021tpami}, Apollo-SouthBay~\cite{lu2019cvpr}, Ford Campus~\cite{Barnes2019icra} and Mulran~\cite{zhang2021pr} datasets. These datasets provide LiDAR scans collected in various environments in multiple countries with the corresponding ground truth poses. 

In particular, we divide the KITTI odometry dataset into the following: sequences 01 and 03-07 for training, sequence 02 for validation, and sequences 00, 08-10 for testing. The other datasets are all used to test the models' generalization capabilities.
We use the two-stage training strategy described in \secref{sec:train}: pre-train the network with dense point matching head in the registration task and then linear probe the global description head for training in the candidate retrieval task.
For the registration task training, we select point cloud pairs with a mix of continuous point cloud pairs with the distance varying from 0-10\,m and close-loop point cloud pairs with an overlap ratio exceeding 0.3~\cite{chen2021rss}.
In the candidate retrieval task training, we use all point cloud pairs with an overlap ratio exceeding 0.3.
We denote the model trained following the above setup as \textit{\name{}} and evaluate its performance on all subsequent tasks.
This model is also utilized for integration with our SLAM system for evaluation.

In addition, to eliminate the impact of different training sets and ensure equitable comparisons with existing baselines, we also train two other models, denoted as \textit{\name{$^\dagger$}} and \textit{\name{$^\diamond$}}, using different dataset splittings that follow the baselines for specific tasks: \textit{\name{$^\dagger$}} is pre-trained on close-loop point cloud pairs, while \textit{\name{$^\diamond$}} is pre-trained on continuous point cloud pairs. Both models are then trained in the candidate retrieval task.
The subsequent experiments will delve into the specific training details for these two models.

\LCTab
\vspace{-0.2cm}
\loopcorrectTab 
\loopcorrectTabp
\vspace{-0.2cm}
\subsection{Candidate Retrieval Performance}
\label{sec:exp_lcd}
To validate the candidate retrieval performance, we follow Chen \etal~\cite{chen2021rss,ma2022ral} and test our approach on the KITTI odometry and Ford Campus dataset.
For a fair comparison, we follow the setup of Chen \etal~\cite{chen2021rss,ma2022ral} and train \textit{\name{$^\dagger$}} on KITTI odomtery sequences 03-10 and validate it on KITTI odomtery sequence 02. Two scans are chosen as a candidate if their overlap value is larger than 0.3.  

\bfvspace{Metrics}
: We use four metrics to evaluate the performance of candidate retrieval: i) the area enclosed by the Receiver Operating Characteristic (ROC) curve and the coordinate axes; ii) Maximum F1 score~(F1max), which is the highest F1 score at different threshold values. 
; iii) Recall@1, which measures the recall when only the most similar candidate frame is selected; iv) Recall@1\%, which measures the recall when the top 1\% of the most similar candidate frames are selected.

\bfvspace{Results}
: The baseline methods used in this experiment are SOTA place recognition methods. For Histogram~\cite{roehling2015iros}, Scan Context~\cite{kim2018iros}, LiDAR-Iris~\cite{wang2020iros}, OverlapNet~\cite{chen2021rss}, PointNetVLAD~\cite{angelina2018cvpr}, NDT-Transformer-P~\cite{zhou2021icra}, MinkLoc3D~\cite{komorowski2021wacv}, and OverlapTranformer~\cite{ma2022ral}, we use the results reported in~\cite{ma2022ral}. For LCDNet~\cite{cattaneo2022tro}, we utilize its official implementation. 
As shown in \tabref{tab:pr}, \name{} and LCDNet achieve cutting-edge performance compared to current advanced methods. However, our method further boosts the baseline by a significant margin for the F1max metric while maintaining a leading position in Recall@1, Recall@1\%, and AUC.
It is worth noting that our global description head does not employ complex designs. The exceptional performance can be attributed to our backbone's outstanding feature representation capabilities and our training strategy, which permits larger batch sizes and more input samples.

\subsection{Closed-Loop Point Cloud Registration Performance}
\label{sec:exp_lc}
We validate the registration performance of our method for closing the loop. 
We follow Cattaneo \etal~\cite{cattaneo2022tro} and test our method on sequences 00 and 08 of the KITTI odometry dataset and sequences 02 and 09 of the KITTI-360 dataset. 
For a fair comparison, we follow Cattaneo \etal to train the \textit{\name{$^\dagger$}} on the KITTI sequence 05-07 and 09, validating it on KITTI sequence 02. The point cloud pairs with ground truth pose distances less than $4$\,m and time intervals greater than $50$\,s are chosen as loop closure samples.

\bfvspace{Metrics}
: In line with~\cite{cattaneo2022tro}, we employ three metrics to evaluate the registration performance at loop closure: i) Relative Translation Error~(RTE), which measures the Euclidean distance between estimated and ground truth translation vectors, ii) Relative Yaw Error~(RYE), which is the average difference between estimated and ground truth yaw angle, and iii) Registration Recall~(RR), representing the fraction of scan pairs with RYE and RTE below certain thresholds, e.g., 5$^\circ$ and 2\,m.

\bfvspace{Results}
: The baseline methods in this experiment include advanced traditional registration methods such as ICP~\cite{besl1992pami} and RANSAC~\cite{fischler1981acm} with FPFH features~\cite{rusu2009icra}. Besides traditional methods, SOTA deep learning approaches are also included, such as RPMNet~\cite{yew2020cvpr}, FCGF~\cite{choy2019iccv}, DGR~\cite{choy2020cvpr}, Predator~\cite{huang2021cvpr}, CofiNet~\cite{yu2021nips}, Geotransformer~\cite{qin2022cvpr}, RDMNet~\cite{shi2023tits} and LCDNet~\cite{cattaneo2022tro}. Furthermore, we also include results from advanced place recognition methods that can output yaw angles, including Scan Context~\cite{kim2018iros}, LiDAR-Iris~\cite{wang2020iros}, and OverlapNet~\cite{chen2021rss}.
For the hand-crafted method, we use the results reported in~\cite{cattaneo2022tro}.
For all DNN-based approaches, we use its official implementation along with open-source models trained on KITTI odometry datasets.
We also report the results of Geotransformer and RDMNet using LGR. LCDNet also provides a fast version using weighted SVD. We denote it as LCDNet~(fast) and report the results.

\tabref{tab:lc2} shows the results on sequences 00 and 08 of the KITTI odometry dataset. As can be seen, our \name{} achieves the best performance across both sequences. \name{} is superior in pose estimation, benefiting from dense, reliable matching. We highlight that its pose estimation accuracy exceeds baselines by a large margin and is comparable to LCDNet refined by ICP.
As illustrated in \tabref{tab:lc2}, the registration recall (RR) for several baselines has reached saturation, reaching $100\%$ in the dataset of the closed loop distance below $4$\,m. To better showcase the superiority of our approach, we have constructed a new test set comprising point cloud pairs with an overlap ratio exceeding $0.3$. The overlap ratio is computed following~\cite{chen2021rss}. These test sets are more challenging, including point cloud pairs with a distance of up to 15\,m. We present the results of several advanced baselines in former tests. As can be seen, our \name{} further amplifies its advantages over other competing approaches. 
Only \name{} maintains 100\% RR and a similar pose estimation accuracy, while others have all declined. 

We highlight that the pose estimation accuracy of \name{} surpasses even LCDNet refined by ICP. Especially regarding translation estimation, \name{} reduces the error by 64\% on seq. 00 and by 16\% on seq. 08. 
This result is encouraging, as current global registration methods typically exhibit inferior registration accuracy compared to geometry-based local registration approaches based on a fine initial guess. As a result, they are commonly employed as initial estimations for methods like the ICP algorithm in practical applications. In this experiment, \name{} demonstrates a noteworthy advancement in registration accuracy compared to the SOTA global registration methods refined by ICP. This significant outcome profoundly underscores the practical advantage of our proposed method in real-world applications.

\tabref{tab:lc3} presents the results on KITTI-360 datasets. As can be seen, the advantages of our method over other baselines are still maintained on KITTI-360 datasets.
Two baselines, i.e., CofiNet and LCDNet, demonstrate comparable registration recall to \name{}. However, both methods fall short compared to \name{} regarding registration accuracy. Additionally, both LCDNet and CofiNet require RANSAC for pose estimation, and LCDNet requires additional ground point filtering. 

From the tests, we have observed that RANSAC-based methods generally outperform RANSAC-free methods.
However, RANSAC is computationally intensive and can account for more than half of the entire registration process, as evident from our runtime experiments as in \secref{sec:exp_runtime}. Nevertheless, our proposed method attains the best performance without relying on RANSAC. By employing a fast solver, LGR, \name{} attains comparable registration performance to that of RANSAC and even surpasses it in terms of translation estimation accuracy. This characteristic offers a significant advantage for integrating our approach within real-time systems.

In sum, our \name{} achieves SOTA performance in closed-loop point cloud registration. It exhibits significant advantages over the baseline methods, offering:
i) The highest level of registration robustness; ii) Exceptional registration accuracy compared to baseline methods, surpassing even the results obtained by baseline with ICP refinement; iii) The ability to achieve best performance without relying on RANSAC.

\registrationTab
\regFig

\vspace{-0.2cm} 	
\subsection{Continuous Relocalization Performance}
\label{sec:exp_con_regis}
To evaluate the relocalization performance, we use LiDAR pairs at most $10$\,m apart as samples, which may cause odometry degeneration.
We test our approach on sequences 08-10 of KITTI odometry, KITTI-360, Apollo-SouthBay, Ford Campus, and Mulran datasets.
For a fair comparison, we follow the setup of prior work~\cite{huang2021cvpr, qin2022cvpr, zhu2022arxiv} and train the model \textit{\name{$^\diamond$}} on KITTI odometry sequence 00-05 and validate it on KITTI odometry sequence 06-07. 
Notably, the sensors, environments, and platform setups differ between the KITTI odometry dataset and other datasets, thoroughly testing the approaches' generalization abilities.

\bfvspace{Metrics}
: We use three metrics to evaluate the registration performance: i) Relative Translation Error~(RTE), which measures the Euclidean distance between estimated and ground truth translation vectors, ii) Relative Rotation Error~(RRE), which measures the geodesic distance between estimated and ground truth rotation matrices, and iii) Registration Recall~(RR), which represents the fraction of scan pairs with RRE and RTE below certain thresholds, e.g., 5$^\circ$ and 2\,m.

\bfvspace{Results}
: We compare the results of our method with the recent RANSAC-based SOTA methods: Predator~\cite{huang2021cvpr}, CofiNet~\cite{yu2021nips}, NgeNet~\cite{zhu2022arxiv}, Geotransformer~\cite{qin2022cvpr}, and RDMNet~\cite{shi2023tits}. We also compare our method using Local-to-Global Registration~(LGR)~\cite{qin2022cvpr} with SOTA RANSAC-free methods: HRegNet~\cite{lu2021iccv}, Geotransformer~\cite{qin2022cvpr} and RDMNet~\cite{shi2023tits}. The results are shown in \tabref{tab:Registration}.

As can be seen, when using RANSAC, our \name{} outperforms all the baselines on all the datasets. 
Especially on the Mulran dataset, RDMNet shows remarkable performance by boosting baselines with a large margin for all three metrics, i.e., increasing registration recall by 10\%, reducing rotation error by 43\%, and reducing translation error by 20\%.
The Mulran dataset poses significant challenges for generalization, as it loses approximately 70$^\circ$ FOV. 
The outstanding performance of our proposed method on the Mulran dataset demonstrates its exceptional generalization capability.
When using LGR, our method attains remarkable results for translation estimation and achieves an average reduction of 35\% in translation error compared to the best results attained through RANSAC-based methods across all datasets.

We conduct a qualitative comparison of registration using \name{} and LCDNet shown in~\figref{fig:reg}. We also present ICP alignment by using the LCDNet prediction as an initial guess to demonstrate the accuracy of \name{}. The first two columns depict the successful cases of both methods. Our \name{} achieves better alignment than LCDNet and LCDNet refined by ICP. The last two columns show the failure cases of LCDNet, while \name{} keeps achieving good alignment.

To provide more insights into the proposed \name{}, we visualize the correspondent points founded by \name{} as in \figref{fig:corr}. The points are aligned using \name{} based on LGR. We observe that there are four characteristics of the regions that the \name{} focuses on: 
i) It effectively captures overlapping regions of two point clouds.
ii) It finds matches over all overlapping regions of two point clouds rather than being limited to a relatively small range, allowing for a wider baseline important for accurate registration.
iii) It neglects the majority of ground points. 
iv) It focuses more on isolated landmarks like tree trunks, signage, vehicles, etc. 
v) It prioritizes important geometric structures like corners and sloping surfaces.

\slamresultFig

\subsection{Study on Runtime}
\label{sec:exp_runtime} 
We report the runtime of \name{} compared to existing SOTA baselines on a system with an Intel i9-10920X CPU and an NVIDIA GTX 3090 GPU. All the methods are evaluated on KITTI sequence 00 with the official implementation.
We present the runtime of point cloud registration in \tabref{tab:runtime_reg} in which the descriptor extraction time also includes the preprocessing required by the respective method. Pairwise Reg. refers to pair registration and uses RANSAC as default.
As depicted, our \name{} using LGR is the fastest method among DNN-based approaches. More commendably, it is also the most robust and accurate one, as shown in \secref{sec:exp_con_regis} and \secref{sec:exp_lc}.

\runtimeldTab
In \tabref{tab:runtime_ld}, the runtime of loop detection is evaluated. 
We emphasize that \name{} exhibits significantly higher efficiency in descriptor extraction when inferring loop detection compared to descriptor extraction during point cloud registration. This efficiency is achieved due to our framework, which only activates the KPEncoder and the global description head during inference, eliminating the need for complex point-level descriptor generation.
In contrast, methods that rely on point-wise descriptors, like LCDNet, have no such substantial runtime reductions for global descriptor extraction.
Regarding map querying, we report the runtime for searching a descriptor in all its previous scans.
For the pairwise comparison, both LiDAR-Iris and OverlapNet employ an ad-hoc comparison function that is hard to optimize in runtime, while \name{} and LCDNet use the Euclidean distance. 
Therefore, we can use the FAISS library~\cite{Johnson2017tbd} for fast searching in \name{} and LCDNet.
We report the querying time of Scan Context using the ring key descriptor for fast searching.
As depicted, our \name{} demonstrates exceptional efficiency in loop retrieval with the support of Faiss.

In summary, \name{} exhibits high efficiency in both registration and candidate retrieval tasks, rendering it well-suited for online SLAM applications.

\subsection{Evaluation of Complete SLAM System}
\label{sec:exp_slam} 
In the previous experiments, we have demonstrated the superior performance and high efficiency of the proposed \name{} in specific subtasks of loop closing and relocalization. We have also verified that the model trained on our dataset splittings, \textit{\name{}}, can achieve comparable SOTA performance with the models specifically trained for each task, \textit{\name{$^\dagger$}} and \textit{\name{$^\diamond$}}. In this section, we will evaluate the performance of our network in solving loop closing and relocalization challenges when integrated with a SLAM system.
We use the SLAM system with A-LOAM as the front-end odometry.
We demonstrate the gradual improvement of SLAM results by integrating our \textit{\name{}} using the LGR estimator for relocalization and loop closing, as depicted in \figref{fig:slam_result}.
The relocalization operates at a frequency of 2.5\,Hz, while loop closing is configured to operate at a frequency of 1\,Hz.

The top row of \figref{fig:slam1} shows the original results of A-LOAM.
As seen, A-LOAM suffers severe degeneration and fails to provide satisfactory results. 
We determine a keyframe's ground truth degeneration flag by comparing the relative pose estimation error to a pre-defined threshold. When the pose error exceeds $1$\,m, the current keyframe is considered degenerated.
The bottom row of \figref{fig:slam1} highlights the degenerated keyframes on the ground truth trajectory in red.
\slamapeFig 
\ablationTab

The bottom row of \figref{fig:slam2} shows the degenerated keyframes detected by our approach. Our approach effectively detects all the degenerations. While this introduces false positives, it avoids missing degeneration detections, which is paramount in practical applications.
The SLAM results are significantly improved by enabling the relocalization module, as in the top row of \figref{fig:slam2}. The trajectory is colored with the distance error between our estimations and ground truth poses. 

The bottom row of \figref{fig:slam3} shows the loop detected and closed by our approach. As in the top row of \figref{fig:slam3}, by enabling both relocalization and loop closing module, the results are further refined compared to \figref{fig:slam2}.

As the code of LCDNet with SLAM is currently unavailable, we compare the results with the most commonly employed loop closing approach, which exploits Scan Context for loop detection and ICP for registration, as in \figref{fig:slam4}.
The bottom row of \figref{fig:slam4} shows the loop detected and closed by Scan Context. As depicted, it introduces several false positives, whereas our approach successfully avoids such occurrences.

We assess different systems' Absolute Position Error (APE) on KITTI odometry sequences 00 and 08. The results of our system and the system employing Scan Context and ICP as the loop closing pipeline are presented in \figref{fig:ape}. By leveraging the flexible structure of the PGO framework, we also substitute the front-end odometry with F-LOAM~\cite{Wang20212021}, thereby demonstrating the capability of our method to integrate diverse odometry systems. As depicted in \figref{fig:ape}, regardless of the chosen odometry front-end, our approach consistently yields more precise localization results than systems utilizing Scan Context and ICP. Moreover, our approach exhibits smaller minimum and maximum errors and enhanced stability.

\subsection{Ablation Study}
\label{sec:exp_ab}

We conduct ablation studies to better understand the effectiveness of each module in our proposed \name{}, as in \tabref{tab:ab}. \tabref{tab:ab_rof} and \tabref{tab:ab_vote} study specific designs of 3D-RoFormer++ and VoteEncoder in continuous registration tasks using Mulran dataset and closed-loop registration task using KITTI dataset~(loop with distance below 4\,m). \tabref{tab:ab_all} and \tabref{tab:ab_train} study overall structure and training strategy in registration task and candidate retrieval task using KITTI and Ford Campus datasets.
We use the model trained on our dataset splittings, \textit{\name{}}, as the default model. The registration performance is evaluated using the RANSAC estimator. 

\bfvspace{Rotary embedding.} \tabref{tab:ab_rof} studies the influence of
different rotary position embedding used in 3D-RoFormer++, where i) nonlinear refers to the embedding using~\eqref{eq:mapping}, ii) linear w/o penalty refers to the embedding using~\eqref{eq:linear_mapping} but trained without penalty loss. 
Using a linear rotary position embedding improves registration robustness and accuracy.
The reason could be attributed to the network's ability to consistently represent relative positions through a linear embedding, thus enhancing the registration performance. 
Using the penalty loss can further help the position embedding learning.

\bfvspace{VoteEncoder.} \tabref{tab:ab_vote} studies the designs of VoteEncoder. Leveraging the central prediction algorithm, VoteEncoder acquires a more robust estimation for the center of the interested region, thus enhancing registration accuracy.
The incorporation of feature aggregation capability via KPConv improves both robustness and accuracy.

\bfvspace{Overall structure.}
\tabref{tab:ab_all} studies the impact of our VoteEncoder and 3D-RoFormer++.
As can be seen, VoteEncoder significantly enhances the registration performance.
The underlying factors contributing to this noteworthy improvement are examined in \secref{sec:exp_whyvote}.
We study the 3D-RoFormer++ by replacing it with a vanilla Transformer using an absolute position encoder~\cite{shi2021ral}. As depicted, 3D-RoFormer++ exhibits superior advantages in registration compared to the vanilla Transformer.

Though neither 3D-RoFormer++ nor VoteEncoder directly contribute to global descriptor generation, they still impact performance.
As illustrated, 3D-RoFormer++ is beneficial in enhancing the capacity of the global descriptor, which suggests that 3D-RoFormer++, compared to vanilla Transformer, is more effective in improving the feature learning of the backbone. On the other hand, using the VoteEncoder does not have a significant impact.
The performance trade-off is justified by the substantial improvement in registration brought about by VoteEncoder.

\bfvspace{Training strategy.} In \tabref{tab:ab_train}, we study the influence of training strategy. We compare the training strategy used in this article, linear probe, with two alternative strategies. The first strategy is fine-tuning, where we also pre-train the model in the registration task and subsequently fine-tune the model along with the global descriptor head in the candidate retrieval task. The second strategy is end-to-end, which starts training the whole network from scratch. We are constrained to smaller batch sizes and fewer positive/negative scans than the linear probe strategy in these two strategies. With the principle of maximizing computational resource utilization, we have set the batch size~(b) to $1$, with $N_p=1$ positive scan~(p) and $N_n=1$ negative scan~(n) for both strategies. As can be seen, the batch size, positive scans,  and negative scans used in these two strategies are significantly less than that used in linear probes, which largely constrains the performance in candidate retrieval. Note that the linear probe also uses the same batch size as $1$ when trained on registration, yet it still achieves the best registration performance. This can be attributed to the linear probe strategy not facing the trade-off between global descriptor capacity and registration capability during training.

\subsection{Insights on How VoteEncoder Contributes to Registration}
\label{sec:exp_whyvote}

In our ablation study, we observed significant performance enhancement brought about by our VoteEncoder in registration tasks. To provide a comprehensive understanding of the benefits of VoteEncoder for registration, we further conducted insight experiments on six additional metrics:
i) Inlier Ratio~(IR), the ratio of correct correspondences with residuals below a certain threshold, e.g., 0.6 m, after applying the ground truth transformation.
ii) Match Recall~(MR), the ratio of true point matches found among all the true matches.
iii) Hit Ratio~(HR), the fraction of true points that have matches among all points that have matches.
iv) Patch Inlier Ratio~(PIR), v) Patch Match Recall~(PMR), and vi) Patch Hit Ratio~(PHR) are with the same meaning as IR, MR, and HR, respectively, but at sparse matching levels. The true patch match represents that patch pairs exhibit actual overlap.
MR and PMR evaluate the coverage ratio of found matches among all true matches, while HR and PHR better assess the match distribution over the overlapping region since a point/patch may possess multiple matches.

We present the result of \name{} and \name{} without VoteEncoder in \tabref{tab:Vote}. To evaluate the efficacy of the VoteEncoder in contrast to an alternative downsampling and aggregation approach, we also present a variant, denoted as \textit{5KPEncoder}, that replaces VoteEncoder with another KPEncoder, comprising $5$ layers of KPEncoder.
As can be seen, using VoteEncoder has shown a decrease in IR and PIR but has notably increased PMR and HR. The rationale behind this can be explained as follows: the VoteEncoder can be seen as a downsampler that filters redundant points and aggregates them into new keypoints. After filtering massive points with similar neighbors, it becomes more difficult to find true matches in sparser candidates~(as observed by the decrease in PIR). However, this also allows \name{} to focus on searching different geometrically salient regions over the point cloud rather than repeatedly sampling the same salient region~(as observed by the increase in PMR and HR). Moreover, this also leads to a more reasonable point patch grouping, making it easier to match an object as a whole and simplifying the subsequent dense matching process (as observed by a decrease in PIR by $17.9$\%, while IR only decreases by $6.3$\%).
As depicted, the significant improvement in PMR and HR brought by VoteEncoder, allowing for wider baselines and more dimensional constraints in registration, greatly reduces registration errors (as observed by a decrease in RRE by 23\% and RTE by 39\%).
Comparison between \textit{5KPEncoder} and full \name{} demonstrates that the VoteEncoder achieves more effective downsampling and better registration than a KPEncoder.

\VoteTab
\corrcompFig
We provide qualitative comparisons of matching points as in \figref{fig:corr_comp}. As depicted, matching points obtained by full \name{} offers enhanced coverage of overlapping regions within the point clouds.
We zoom in and examine four specific local areas: car, tree, signage, and shrub within both scenes. It is evident that the full \name{} effectively groups and matches points of local areas.

\bfvspace{Insights}. The above results lead us to rethink which indirect metrics are more important in point cloud registration. Apart from the final performance evaluated by direct metrics of Registration Recall~(RR), Relative Rotation Error~(RRE), and Relative Translation Error~(RTE), most prior work~\cite{shi2023tits,choy2019iccv,bai2020cvpr,yew2018eccv,deng2018cvpr,deng2018eccv,li2019iccv,qin2022cvpr,zhu2022arxiv,yu2021nips,huang2021cvpr} focuses only on the keypoint salience evaluated by indirect metric Inlier Ratio~(IR), also referred to as precision. However, higher IR does not guarantee higher robustness or accuracy, as observed in \tabref{tab:Vote}. Conversely, the coverage ratio of found matches among all true matches, as evaluated by metrics MR and HR, should be a crucial factor to consider. These metrics provide insights into the distribution of matching points and can guide the design of new keypoint detection methods by considering this distribution. It is important to note that the coverage ratio is just one minor aspect within the broader research context of distribution, and there are still numerous other factors, such as divergence and distribution across dimensions, waiting to be explored in this field.

Indeed, numerous existing keypoint detection and matching methods~\cite{bai2020cvpr,li2019iccv} have proposed to prevent keypoints from being excessively concentrated in local areas. Earlier work for LiDAR odometry, LOAM~\cite{zhang2014rss} and its variants~\cite{Shan2018iros,Wang20212021} propose to extract keypoints uniformly from all LiDAR scan lines. However, they have not provided a clear explanation for the underlying reasons.
We are the first to specifically propose this insight and validate it through experimental cases by providing quantitative metrics.

\section{Conclusion}
\label{sec:conclusion}
This article analyzes the challenges and commonalities in loop closing and relocalization tasks and proposes a unified framework to address both tasks. Based on this framework, we present \name{}, a novel multi-head network for candidate retrieving and point cloud registration.
\name{} incorporates a novel keypoint detection module, offering three types of features to meet different requirements for distinct post-processing. Two novel sub-modules, 3D-RoFormer++ and VoteEncoder, are proposed for feature generation. 3D-RoFormer++ learns contextual information between two point clouds and leverages a novel transformation-invariant rotary position embedding for geometric information aggregation. VoteEncoder, on the other hand, learns to detect keypoints from distinct objects in the scene while maintaining a uniform, non-repeating distribution. 
Both 3D-RoFormer++ and VoteEncoder significantly contribute to our network's loop closing and relocalization performance. 
Given the impressive improvement in registration performance brought by VoteEncoder, we conduct experiments and evaluation metrics to analyze the underlying reasons and obtain an important insight that point matching should consider distribution rather than solely focusing on precision.
We demonstrate the SOTA performance of our \name{} in three sub-tasks related to loop closing and relocalization. It is shown that the best performance of \name{} does not rely on a costly robust estimator and outperforms the baseline method refined by ICP.
Finally, we integrate our network as the loop closing and relocalization module in a complete SLAM system and verify its capability to address loop closing and relocalization tasks.

\small{
\bibliographystyle{ieeetran}
\bibliography{reference,glorified}
}

\end{document}

%% file: stachnisslab-latex.tex
\makeatletter
\renewcommand{\maketag@@@}[1]{\hbox{\m@th\normalsize\normalfont#1}}%
\makeatother

\usepackage[font=small]{caption}

\def\secref#1{Sec.~\ref{#1}}
\def\figref#1{Fig.~\ref{#1}}
\def\tabref#1{Tab.~\ref{#1}}
\def\eqref#1{Eq.~(\ref{#1})}
\def\algref#1{Alg.~\ref{#1}}

\makeatletter
\usepackage{xspace}
\DeclareRobustCommand\onedot{\futurelet\@let@token\@onedot}
\def\@onedot{\ifx\@let@token.\else.\null\fi\xspace}

\def\etal{{et al}\onedot}
\makeatother

\usepackage{array}
\newcolumntype{L}[1]{>{\raggedright\let\newline\\\arraybackslash\hspace{0pt}}m{#1}}
\newcolumntype{C}[1]{>{\centering\let\newline\\\arraybackslash\hspace{0pt}}m{#1}}
\newcolumntype{R}[1]{>{\raggedleft\let\newline\\\arraybackslash\hspace{0pt}}m{#1}}

%% file: stachnisslab-math.tex
\renewcommand{\d}[1]{\b {#1}}

\newcommand{\m}[1]{{\mbox{{\sffamily\slshape{#1\/}}}}}
\newcommand{\mi}[1]{{\mbox{{\fontencoding{T1}\sffamily\slshape{\scriptsize#1\/}}}}}	
\newcommand{\mii}[1]{{\mbox{{\fontencoding{T1}\sffamily\slshape{\tiny#1\/}}}}}	

\newcommand{\trans}[0]{^{\sf T}}            



%% file: table.tex
\usepackage{rotating}

\newcommand{\LCTab}{
	\begin{table}
		
		\caption{Candidate Retrieval Results on KITTI and Ford Campus.}
		\centering
		\setlength{\tabcolsep}{5pt}
		\renewcommand\arraystretch{1}
		\begin{tabular}{l|l|cccc}
			
			\toprule
			
			\multirow{2}*{Dataset}&\multirow{2}*{Method}&\multirow{2}*{AUC}    &\multirow{2}*{F1max}  & {Recall} 		&Recall \\
			&&   & & @1  		&@1\% \\

			\midrule
			\multirow{10}*{KITTI}&Histogram~\cite{roehling2015iros}	
			&0.826 		&0.825 	&0.738 	&0.871	\\
			
			&Scan Context~\cite{kim2018iros}		
			&0.836&0.835&0.820&0.869				\\
			
			&LiDAR-Iris~\cite{wang2020iros}		
			&0.843&0.848&0.835&0.877\\
			
			&PointNetVLAD~\cite{angelina2018cvpr}
			&0.856&0.846&0.776&0.845\\
			
			&OverlapNet~\cite{chen2021rss}		
			&0.867&0.865&0.816&0.908	\\
			
			&NDT-Transformer-P~\cite{zhou2021icra}	
			&0.855&0.856&0.802&0.869\\
			
			&MinkLoc3D~\cite{komorowski2021wacv}	
			&0.894&0.869&0.876&0.920 \\
			
			&OverlapTransformer~\cite{ma2022ral}	
			&{0.907}&{0.877}&{0.906}&{0.964} \\
			
			&LCDNet	
			& \underline{0.933} &\underline{0.883} &\underline{0.915} &\underline{0.974}   \\
			
			&\name$^\dagger$
			& \textbf{0.945} &\textbf{0.907} &\textbf{0.926} &\textbf{0.980}   \\
			
			&\textg{\name}	
			&\textg{0.958} &\textg{0.922} &\textg{0.937} &\textg{0.993}  \\

			\bottomrule	
			\multirow{10}*{\makecell[l]{Ford\\ Campus}}&Histogram~\cite{roehling2015iros}	
			&0.841 		&0.800 	&0.812 	&0.897	\\
			
			&Scan Context~\cite{kim2018iros}		
			&0.903&0.842&0.878&0.958				\\
			
			&LiDAR-Iris~\cite{wang2020iros}		
			&0.907&0.842&0.849&0.937\\
			
			&PointNetVLAD~\cite{angelina2018cvpr}
			&0.872&0.830&0.862&0.938\\
			
			&OverlapNet~\cite{chen2021rss}		
			&0.854&0.843&0.857&0.932	\\
			
			&NDT-Transformer-P~\cite{zhou2021icra}	
			&0.835&0.850&0.900&0.927\\
			
			&MinkLoc3D~\cite{komorowski2021wacv}	
			&0.871&0.851&0.878&0.942 \\
			
			&OverlapTransformer~\cite{ma2022ral}	
			&{0.923}&{0.856}&{0.914}&{0.954} \\
			
			&LCDNet	
			& \underline{0.961} &\underline{0.908} &\underline{0.949} &\underline{0.984}   \\
			
			&\name$^\dagger$	
			& \textbf{0.974} &\textbf{0.929} &\textbf{0.951} &\textbf{0.985}   \\
			
			&\textg{\name}	
			&\textg{0.972} &\textg{0.920} &\textg{0.932} &\textg{0.987} \\
			
			\bottomrule

		\end{tabular}
		\begin{tablenotes} 
			\item The best results are highlighted in bold, and the second best in underlines.
		\end{tablenotes} 
		\label{tab:pr}
		
		\vspace{-0.4cm}
\end{table}}

\newcommand{\loopcorrectTab}{
	\begin{table*}
		
		\caption{Point Cloud Registration Results on Closed Loop of KITTI Odometry Dataset.}
		\centering
		\setlength{\tabcolsep}{6pt}
		\renewcommand\arraystretch{1}
		
		\begin{tabular}{cl|ccc|ccc}
			\toprule
			\multicolumn{8}{c}{\textit{Closed loop with distance below $4$ m}}\\
			\midrule
			&&\multicolumn{3}{c|}{Seq. 00}     &\multicolumn{3}{c}{Seq. 08}   \\
			
			&&RR(\%)&RTE(m)[succ./all]&RYE($^\circ$)[succ./all]     &RR&RTE(m)[succ./all]&RYE($^\circ$)[succ./all]    	\\
			
			\midrule
			\multirow{9}*{\rotatebox{90}{RANSAC-based}}
			
			&RANSAC~\cite{fischler1981acm}		
			&{33.95}&{0.98/2.75}&{1.37/12.01} 		&15.61 &1.33/4.57&1.79/37.31			\\
			
			&FCGF~\cite{choy2019iccv}		
			&{47.31}&{1.05/2.07}&{0.60/1.88} 		&27.80 &1.28/2.42&2.38/159.48			\\
			
			&Predator~\cite{huang2021cvpr}		
			&\underline{98.93}&{0.07/0.13}&{0.11/0.45} 		&{99.70} &0.10/0.11&{0.34/0.53}\\
			
			&CofiNet~\cite{yu2021nips}		
			&\textbf{100}&\underline{0.07/0.07}&\underline{0.10/0.10} 		&\textbf{100} &\underline{0.10/0.10}&\textbf{0.33/0.33}\\
			
			&{Geotransformer~\cite{qin2022cvpr}	}	
			&{98.11}&{0.09/0.28}&{0.11/0.13} 		&{99.12} &0.11/0.18&{0.35/1.76}\\
			
			&RDMNet~\cite{shi2023tits}		
			&{98.36}&{0.07/0.27}&{0.11/0.28} 		&\underline{99.80} &0.10/0.14&{0.34/0.49}\\
			
			&LCDNet~\cite{cattaneo2022tro}		
			&\textbf{100}&{0.11/0.11}&{0.12/0.12} 		&\textbf{100} &0.15/0.15&\underline{0.34/0.34}\\
			
			&\name{$^\dagger$}
			&\textbf{100}&\textbf{0.04/0.04}&\textbf{0.09/0.09}     &\textbf{100}&\textbf{0.08/0.08}&\textbf{0.33/0.33}   \\
			
			&\textg{\name}	
			&\textg{100}&\textg{0.04/0.04}&\textg{0.09/0.09}     &\textg{100}&\textg{0.08/0.08}&\textg{0.33/0.33}   \\
			\midrule
			\multirow{11}*{\rotatebox{90}{RANSAC-free}}
			
			&Scan Context*~\cite{kim2018iros}	
			&{97.66}&{-/-}&{1.34/1.92} 		&98.21 &-/-&1.71/3.11			\\
			
			&LiDAR-Iris*~\cite{wang2020iros}		
			&\underline{98.83}&{-/-}&{0.65/1.69} 		&99.29 &-/-&0.93/1.84			\\
			
			&OverlapNet*~\cite{chen2021rss}		
			&{83.86}&{-/-}&{1.28/3.89} 		&0.10 &-/-&2.03/65.45			\\
			
			&ICP~\cite{besl1992pami}
			&{35.57}&{0.97/2.08}&{1.36/8.98} 		&0 &-/2.43&-/160.46			\\
			
			&DGR~\cite{choy2020cvpr}		
			&{95.11}&{0.57/0.74}&{0.41/2.70} 		&2.42 &1.13/7.72&3.13/135.53			\\
			
			&RPMNet~\cite{yew2020cvpr}		
			&{47.31}&{1.05/2.07}&{0.60/1.88} 		&27.80 &1.28/2.42&1.77/13.13			\\
			
			&LCDNet~(fast)~\cite{cattaneo2022tro}		
			&{93.03}&{0.65/0.77}&{0.86/1.07} 		&{60.71} &{1.02/1.62}&{1.65/3.13}\\
			
			&{Geotransformer~(LGR)~\cite{qin2022cvpr}	}	
			&{96.72}&{0.10/0.41}&{0.12/0.99} 		&{97.06} &0.16/0.35&{0.46/2.85}\\
			
			&RDMNet~(LGR)~\cite{shi2023tits}		
			&{97.57}&\underline{0.06/0.37}&{0.12/0.64} 		&\underline{99.36} &\underline{0.10/0.22}&\underline{0.34/0.59}\\
			
			&\name$^\dagger$~(LGR)
			&\textbf{100}&\textbf{0.04/0.04}&\textbf{0.09/0.09}     &\textbf{100}&\textbf{0.08/0.08}&\textbf{0.33/0.33}   \\
			
			&\textg{\name~(LGR)}	
			&\textg{100}&\textg{0.04/0.04}&\textg{0.09/0.09}     &\textg{100}&\textg{0.07/0.07}&\textg{0.33/0.33}   \\
			\midrule
			&{LCDNet+ICP~\cite{cattaneo2022tro}		}
			&{100}&{0.04/0.04}&{0.09/0.09} 		&{100} &{0.09/0.09}&{0.33/0.33}\\
			
		\end{tabular}

		\renewcommand\arraystretch{1}
		\begin{tabular}{cl|ccc|ccc}
			\toprule
			\multicolumn{8}{c}{\textit{Closed loop with overlap beyond $0.3$}}\\
			\midrule
			
			&&\multicolumn{3}{c|}{Seq. 00}     &\multicolumn{3}{c}{Seq. 08}   \\
			
			&&RR(\%)&RTE(m)[succ./all]&RYE($^\circ$)[succ./all]     &RR&RTE(m)[succ./all]&RYE($^\circ$)[succ./all]    	\\
			
			\midrule
			\multirow{7}*{\rotatebox{90}{RANSAC-based}}
			&Predator~\cite{huang2021cvpr}		
			&{80.52}&{0.10/2.13}&{0.14/13.29} 		&{81.98} &0.15/1.93&{0.37/23.05}\\
			
			&CofiNet~\cite{yu2021nips}		
			&\underline{98.21}&\underline{0.09/0.29}&\underline{0.13/1.13} 		&\underline{99.05} &\underline{0.13/0.24}&\underline{0.35/1.49}\\
			
			&Geotransformer~\cite{qin2022cvpr}		
			&{82.07}&{0.10/3.04}&{0.13/4.53} 		&{92.69} &0.15/0.97&{0.41/10.29}\\
			
			&RDMNet~\cite{shi2023tits}		
			&{71.36}&{0.13/4.53}&{0.23/7.81} 		&{92.94} &0.16/1.37&{0.46/6.09}\\
			
			&LCDNet~\cite{cattaneo2022tro}		
			&{94.86}&{0.23/0.89}&{0.24/3.93} 		&{96.87} &0.31/0.62&{0.51/1.74}\\

			&\name$^\dagger$	
			&\textbf{100}&\textbf{0.05/0.05}&\textbf{0.10/0.10}     &\textbf{100}&\textbf{0.08/0.08}&\textbf{0.34/0.34}   \\
			
			&\textg{\name}	
			&\textg{100}&\textg{0.05/0.05}&\textg{0.10/0.10}     &\textg{100}&\textg{0.08/0.08}&\textg{0.34/0.34}   \\
			
			\midrule
			\multirow{5}*{\rotatebox{90}{\scriptsize{RANSAC-free}}}
			
			&LCDNet~(fast)~\cite{cattaneo2022tro}		
			&{66.82}&{0.56/2.02}&{0.68/6.12} 		&{47.29} &{0.88/3.00}&{1.06/5.28}\\
			
			&Geotransformer~(LGR)~\cite{qin2022cvpr}		
			&\underline{80.43}&\underline{0.11/3.33}&\underline{0.16/7.91} 		&{87.85} &{0.20/1.47}&{0.56/12.78}\\
			
			&RDMNet~(LGR)~\cite{shi2023tits}		
			&{70.19}&{0.13/5.06}&{0.23/11.14} 		&\underline{91.62} &\underline{0.15/1.98}&\underline{0.48/7.16}\\
			
			&\name$^\dagger$~(LGR)
			&\textbf{100}&\textbf{0.04/0.04}&\textbf{0.10/0.10}     &\textbf{100}&\textbf{0.08/0.08}&\textbf{0.33/0.33}   \\
			
			&\textg{\name~(LGR)}	
			&\textg{100}&\textg{0.04/0.04}&\textg{0.10/0.10}     &\textg{100}&\textg{0.08/0.08}&\textg{0.34/0.34}   \\
			\midrule
			&{LCDNet+ICP}~\cite{cattaneo2022tro}		
			&{94.98} &{0.11/0.77}&{0.14/3.84} 		&{96.98} &{0.18/0.48}&{0.43/1.68}\\

			\bottomrule

		\end{tabular}
		\begin{tablenotes} 
			\item Superscript * means the approaches only estimate yaw angle.  The best results are highlighted in bold, and the second best in underlines.
			\item 
		\end{tablenotes} 
		\label{tab:lc2}
		
		\vspace{-0.4cm}
\end{table*}}

\newcommand{\loopcorrectTabp}{
	\begin{table*}
		
		\caption{Point Cloud Registration Results on Closed Loop of KITTI360.}
		\centering
		\setlength{\tabcolsep}{6pt}
		\renewcommand\arraystretch{1}
		
		\begin{tabular}{cl|ccc|ccc}
			
			\toprule
			
			&&\multicolumn{3}{c|}{Seq. 02}     &\multicolumn{3}{c}{Seq. 09}   \\

			&&RR(\%)&RTE(m)[succ./all]&RYE($^\circ$)[succ./all]     &RR&RTE(m)[succ./all]&RYE($^\circ$)[succ./all]    	\\
			
			\midrule
			\multirow{9}*{\rotatebox{90}{RANSAC-based}}
			
			&RANSAC~\cite{fischler1981acm}		
			&{24.78}&{1.24/3.67}&{1.83/32.22} 		&29.69 &1.12/3.14&1.48/23.42			\\
			
			&FCGF~\cite{choy2019iccv}		
			&{12.43}&{0.56/11.32}&{0.62/146.77} 		&62.40 &0.47/5.06&0.44/60.64			\\
			
			&Predator~\cite{huang2021cvpr}		
			&{97.56}&\underline{0.22/0.34}&\underline{0.28/1.19} 		&\underline{98.44} &\underline{0.14/0.22}&{0.18/0.66}\\
			
			&CofiNet~\cite{yu2021nips}		
			&{98.56}&{0.25/0.30}&{0.30/0.35} 		&\textbf{100} &0.15/0.15  &\underline{0.19/0.19}\\
			
			&Geotransformer~\cite{qin2022cvpr}		
			&{92.32}&{0.26/0.84}&{0.49/9.71} 		&{96.24} &0.16/0.50 &{0.25/1.84}\\

			&RDMNet~\cite{shi2023tits}		
			&{94.98}&{0.23/0.66}&{0.38/0.97} 		&{83.27} &0.19/2.24 &{0.31/3.04}\\
			
			&LCDNet~\cite{cattaneo2022tro}		
			&\underline{98.62}&{0.28/0.32}&{0.32/0.35} 		&\textbf{100} &0.18/0.18&{0.20/0.20}\\

			&\name$^\dagger$	
			&\textbf{98.63}&\textbf{0.21/0.26}&\textbf{0.27/0.30}     &\textbf{100}&\textbf{0.11/0.11}&\textbf{0.16/0.16}   \\
			
			&\textg{\name}	
			&\textg{98.64}&\textg{0.21/0.26}&\textg{0.27/0.30}     &\textg{100}&\textg{0.11/0.11}&\textg{0.16/0.16}   \\
			\midrule
			\multirow{11}*{\rotatebox{90}{RANSAC-free}}
			
			&Scan Context*~\cite{kim2018iros}	
			&{92.31}&{-/-}&{1.60/5.49} 		&\underline{95.29} &-/-&1.40/6.80			\\
			
			&LiDAR-Iris*~\cite{wang2020iros}		
			&\underline{96.54}&{-/-}&{1.71/3.44} 		&86.26 &-/-&1.51/7.08			\\
			
			&OverlapNet*~\cite{chen2021rss}		
			&{11.42}&{-/-}&{1.79/76.74} 		&54.33 &-/-&1.38/33.62			\\
			
			&ICP~\cite{besl1992pami}
			&{4.19}&{1.10/2.26}&{1.74/149.76} 		&21.24 &1.06/2.22&1.34/66.34			\\
			
			&DGR~\cite{choy2020cvpr}		
			&{14.26}&{0.66/6.68}&{0.72/126.81} 		&63.51 &0.50/3.14&0.38/53.83			\\
			
			&RPMNet~\cite{yew2020cvpr}		
			&{37.99}&{1.18/2.26}&{1.30/5.97} 		&41.42 &1.13/2.21&1.02/3.95			\\
			
			&LCDNet~(fast)~\cite{cattaneo2022tro}		
			&{82.92}&{0.84/1.10}&{1.28/1.67} 		&{89.49} &{0.76/0.94}&{0.99/1.19}\\
			
			&Geotransformer~(LGR)~\cite{qin2022cvpr}		
			&{91.24}&{0.25/1.01}&{0.61/7.90} 		&{94.23} &\underline{0.17/0.70} &\underline{0.29/2.70}\\
			
			&RDMNet~(LGR)~\cite{shi2023tits}		
			&{94.44} &\underline{0.19/0.72} &\underline{0.36/1.43} 		&{80.36} &{0.18/2.82} &{0.31/5.64}\\
			
			&\name$^\dagger$~(LGR)
			&\textbf{98.63}&\textbf{0.16/0.21}&\textbf{0.24/0.27}     &\textbf{100}&\textbf{0.10/0.10}&\textbf{0.14/0.14}   \\
			
			&\textg{\name~(LGR)}	
			&\textg{98.59}&\textg{0.16/0.21}&\textg{0.23/0.26}     &\textg{100}&\textg{0.09/0.09}&\textg{0.14/0.14}   \\
			\midrule
			&{LCDNet+ICP~\cite{cattaneo2022tro}	}	
			&{98.51}&{0.20/0.25}&{0.24/0.27} 		&{100} &{0.10/0.10}&{0.15/0.15}\\

			\bottomrule	
			
		\end{tabular}
		\begin{tablenotes} 
			\item Superscript * means the approaches only estimate yaw angle. The best results are highlighted in bold, and the second best in underlines.
		\end{tablenotes} 
		
		\label{tab:lc3}
		
		\vspace{-0.4cm}
\end{table*}}

\newcommand{\registrationTab}{
	\begin{table}
		\caption{Continuous Registration Results on Multiple Datasets. All the Models Are Only Trained on the KITTI Dataset.}
		\centering
		\setlength{\tabcolsep}{2.5pt}
		\renewcommand\arraystretch{1}
		
		\begin{tabular}{cl|c|c|c|c|c}
			\toprule
			&&KITTI     &KITTI-360   & Apollo  		&Ford 	&Mulran\\
			
			\midrule
			\multicolumn{7}{c}{\textit{Registration Recall}~(\%)}\\
			\midrule
			\multirow{7}*{\rotatebox{90}{\scriptsize{RANSAC-based}}}
			&Predator~\cite{huang2021cvpr}	
			&\textbf{99.82} 		&99.50 			&{99.27}  	&{99.32} 	&53.02\\ 
			
			&CofiNet~\cite{yu2021nips}		
			&\textbf{99.82}			&99.62 			&\textbf{100}  		&{99.32} 		&80.79 \\ 
			
			&NgeNet~\cite{zhu2022arxiv}		
			&\textbf{99.82}	    	&\textbf{99.94} 	&\textbf{100}  		&\textbf{100} &{82.96}\\ 
			
			&Geotransformer~\cite{qin2022cvpr}
			&\textbf{99.82} 		&99.86 		&\textbf{100}  		&\textbf{100} &75.68\\
			
			&RDMNet~\cite{shi2023tits}	
			&\textbf{99.82}  &{99.89}  &\textbf{100} &\textbf{100} &{87.09}  \\
			
			&\name$^\diamond$	
			&\textbf{99.82}  &\textbf{99.94}  &\textbf{100} &\textbf{100} &\textbf{97.89}  \\
			
			&\textg{\name}	
			&\textg{99.82} &\textg{99.94} &\textg{100} &\textg{100} &\textg{98.22}  \\
			
			\midrule
			\multirow{5}*{\rotatebox{90}{\scriptsize{RANSAC-free}}}
			
			&HRegNet~\cite{lu2021iccv}	
			&96.76 &20.39		&9.39		&90.48 &-\\
			
			&Geotransformer~(LGR)~\cite{qin2022cvpr}	
			&\textbf{99.82} &99.86 		&\textbf{100}  		&98.64 &72.91\\
			
			&RDMNet~(LGR)~\cite{shi2023tits}	
			& \textbf{99.82} &{99.90} &\textbf{100} &\textbf{100} &{83.68}  \\
			
			&\name$^\diamond$~(LGR)	
			&\textbf{99.82} &\textbf{99.94} &\textbf{100} &\textbf{100} &\textbf{96.76}  \\
			
			&\textg{\name~(LGR)}	
			&\textg{99.82} &\textg{99.94} &\textg{100} &\textg{100} &\textg{97.53}  \\

			\midrule
			\multicolumn{7}{c}{\textit{Relative Rotation Error}~($^\circ$)}\\
			\midrule
			\multirow{7}*{\rotatebox{90}{\scriptsize{RANSAC-based}}}
			&Predator~\cite{huang2021cvpr}	
			&0.25 		&0.29 			&0.21  		&0.38 	&1.03\\ 
			
			&CofiNet~\cite{yu2021nips}		
			&0.37		&0.44 			&\underline{0.18}  		&0.37 	&0.52\\ 
			
			&NgeNet~\cite{zhu2022arxiv}		
			&0.26	    &0.30 			&\underline{0.18} 		&{0.26} 	&0.35\\ 
			
			&Geotransformer~\cite{qin2022cvpr}
			&{0.22} 	&{0.28}	&{0.28}  		&{0.28} 	&\underline{0.30}\\
			
			&RDMNet~\cite{shi2023tits}
			&\underline{0.18} &\underline{0.25} &\textbf{0.10} &\underline{0.21} 	&{0.45}  \\
			
			&\name$^\diamond$
			&\textbf{0.17} &\textbf{0.22} &\textbf{0.10} &\textbf{0.18} 	&\textbf{0.17}  \\
			
			&\textg{\name}	
			&\textg{0.19} &\textg{0.24} &\textg{0.09} &\textg{0.16} &\textg{0.17}  \\
			
			\midrule
			\multirow{5}*{\rotatebox{90}{\scriptsize{RANSAC-free}}}
			
			&HRegNet~\cite{lu2021iccv}	
			&1.04 &2.18		&2.14		&0.91 &-\\
			
			&Geotransformer~(LGR)~\cite{qin2022cvpr}	
			&0.31 	&0.36 		&0.29  		&0.50 	&0.49\\
			
			&RDMNet~(LGR)~\cite{shi2023tits}
			&\textbf{0.27} &\textbf{0.35} &\textbf{0.29} &\textbf{0.39} &\underline{0.48}  \\
			
			&\name$^\diamond$~(LGR)	
			&\underline{0.35} &\underline{0.36} &\underline{0.30} &\underline{0.42} &\textbf{0.45}  \\
			
			&\textg{\name~(LGR)}	
			&\textg{0.35} &\textg{0.36} &\textg{0.29} &\textg{0.34} &\textg{0.43}  \\

			\midrule
			\multicolumn{7}{c}{\textit{Relative Translation Error}~(cm)}\\
			\midrule
			\multirow{7}*{\rotatebox{90}{\scriptsize{RANSAC-based}}}
			&Predator~\cite{huang2021cvpr}	
			&{5.8}		&{7.2} 			&7.8  		&11.7 	&30.04\\ 
			
			&CofiNet~\cite{yu2021nips}		
			&8.2		&10.1 			&6.7  		&11.6 	&17.3\\ 
			
			&NgeNet~\cite{zhu2022arxiv}		
			&6.1    	&7.5			&{5.9}  		&{7.9}   &\underline{9.2}\\ 
			
			&Geotransformer~\cite{qin2022cvpr}
			&6.7		&8.1 			&{6.1}  		&{10.6} 	&12.0\\
			
			&RDMNet~\cite{shi2023tits}
			&\underline{5.3} &\underline{7.0} &\underline{4.6} &\underline{7.6} &{14.4}\\
			
			&\name$^\diamond$
			&\textbf{3.9} &\textbf{5.7} &\textbf{3.6} &\textbf{7.1} 	&\textbf{7.5}  \\
			
			&\textg{\name}	
			&\textg{3.9} &\textg{5.8} &\textg{3.4} &\textg{6.6} &\textg{7.4}  \\
			
			\midrule
			\multirow{5}*{\rotatebox{90}{\scriptsize{RANSAC-free}}}
			
			&HRegNet~\cite{lu2021iccv}	
			&6.7 &112.8		&116.7		&16.5 &-\\
			
			&Geotransformer~(LGR)~\cite{qin2022cvpr}
			&5.5 &{6.9} 	&5.1  		&\underline{12.2} 	&9.7\\
			
			&RDMNet~(LGR)~\cite{shi2023tits}	
			&\underline{3.9} &\underline{5.9} &\underline{3.5} &\textbf{6.1} &\underline{9.3}  \\
			
			&\name$^\diamond$~(LGR)	
			&\textbf{2.7} &\textbf{5.0} &\textbf{2.5} &\textbf{6.1} &\textbf{6.2}  \\
			
			&\textg{\name~(LGR)}	
			&\textg{2.8} &\textg{5.4} &\textg{2.5} &\textg{6.5} &\textg{6.2}  \\
			
			\bottomrule	
		\end{tabular}
		\begin{tablenotes} 
			\item The best results are highlighted in bold, and the second best in underlines.
		\end{tablenotes} 
		\vspace{-0.4cm}
		\label{tab:Registration}
\end{table}}

\newcommand{\runtimeldTab}{
	\begin{table}
		\caption{Comparison of Inference Time.}
		\centering
		\renewcommand\arraystretch{1}
		\vspace{-0.2cm}
		
		\begin{subtable}{0.95\linewidth}
			\caption{Inference time for point cloud registration.}
			\vspace{-0.2cm}
			\label{tab:runtime_reg}
			\centering
			\setlength\tabcolsep{1.5pt}
			\begin{tabular}{cl|ccc}
				\toprule
				&\multirow{2}*{Model} &{Descriptor } & Pairwise  & \multirow{2}*{Total (ms)}\\
				&&Extraction (ms) &Reg. (ms)  &\\
				\midrule
				\multirow{2}*{\rotatebox{90}{Hand}}
				&RANSAC		
				&{135} &156 &291\\
				&FGR
				&135   &246 &381\\
				
				\midrule
				\multirow{6}*{\rotatebox{90}{DNN-based}}
				
				&Predator
				&173 	&244  &417 \\
				&CofiNet
				&{377} 	&257  &634 \\
				&LCDNet
				&{235} &431	 &666 \\
				&LCDNet+ICP
				&{235} &566	 &801 \\
				&\name{} 	
				&{{293}} &{390}  	 & {{683}}	\\
				&\name{} (LGR)
				&{{293}} &{13}  	 & {{306}}	\\
				\bottomrule	
				
			\end{tabular}
		\end{subtable}
		
		\begin{subtable}{0.95\linewidth}
			\caption{Inference time for candidate retrieval.}
			\vspace{-0.2cm}
			\label{tab:runtime_ld}
			\centering
			\setlength\tabcolsep{2pt}
			\begin{tabular}{cl|ccc}
				\toprule
				&\multirow{2}*{Model} &{Descriptor } & Pairwise  & Map \\
				&&Extraction (ms) &Comp. (ms)  &querying (ms)\\
				\midrule
				\multirow{2}*{\rotatebox{90}{Hand}}
				&Scan Context		
				&{1} &0.09 &6\\
				&LiDAR-Iris
				&10 		&9  		 &42037\\
				
				\midrule
				\multirow{3}*{\rotatebox{90}{DNN}}
				&OverlapNet
				&15 		&6  		 &26634\\
				&LCDNet
				&{182} 		&0.01	 &5 \\
				&\name{} 	
				&{50}       &{0.01}  	 & {5}	\\
				\bottomrule	
				
			\end{tabular}
		\end{subtable}
		\vspace{-0.4cm}
\end{table}}

\newcommand{\ablationTab}{
	\begin{table*}
		\caption{Ablation Studies of Individual Modules.}
		
		\label{tab:ab}
		\centering
		\setlength\tabcolsep{2pt}
		\renewcommand\arraystretch{1}
		
		\begin{subtable}{0.48\linewidth}
			\caption{\textbf{Rotary embedding in 3D-RoFormer++}. Linear embedding is effective.}\label{tab:ab_rof}
			\vspace{-0.1cm}
			\centering
			\begin{tabular}{l|ccc|ccc}
				\toprule
				\multirow{2}*{case} &\multicolumn{3}{c|}{Mulran}  		&\multicolumn{3}{c}{KITTI-loop}\\
				&\multirow{1}*{RR(\%)}  &\multirow{1}*{RRE($^\circ$)}  &\multirow{1}*{RTE(cm)} &\multirow{1}*{RR(\%)}  &\multirow{1}*{RRE($^\circ$)}  &\multirow{1}*{RTE(cm)}	\\
				
				\midrule	
				nonlinear &94.97 		&0.18 			&8.1  	
				&99.86 	   	&0.29 &5.9\\ 
				\rowcolor{gray!30}
				{linear} &\textbf{98.22} &\textbf{0.17} &\textbf{7.4} 
				&\textbf{100} &\textbf{0.21} &\textbf{4.3}\\
				linear w/o penalty  &94.29 &0.18&7.5 		
				&\textbf{100} 	&{0.23}  &\textbf{4.3}\\
				\bottomrule	
				
			\end{tabular}
		\end{subtable}
		\vspace{0.1cm}
		\begin{subtable}{0.48\linewidth}
			\caption{\textbf{VoteEncoder}. The full VoteEncoder performs the best.}\label{tab:ab_vote}
			\vspace{-0.1cm}
			\centering
			\begin{tabular}{l|ccc|ccc}
				\toprule
				\multirow{2}*{case} &\multicolumn{3}{c|}{Mulran}  		&\multicolumn{3}{c}{KITTI-loop}\\
				&\multirow{1}*{RR(\%)}  &\multirow{1}*{RRE($^\circ$)}  &\multirow{1}*{RTE(cm)} &\multirow{1}*{RR(\%)}  &\multirow{1}*{RRE($^\circ$)}  &\multirow{1}*{RTE(cm)}	\\
				
				\midrule	\rowcolor{gray!30}
				full &\textbf{98.22} &\textbf{0.17} &\textbf{7.4} 
				&\textbf{100} &\textbf{0.21} &\textbf{4.3}\\
				
				KPConv w/o &93.40	 	&0.18			&8.0		&99.97&0.26&5.0\\
				
				central prediction w/o &92.80 &0.19&8.0 		&{99.99} 	&0.24  &4.5\\
				\bottomrule

			\end{tabular}
		\end{subtable}

		\begin{subtable}{0.99\linewidth}
			\caption{\textbf{Overall structure}. Both VoteEncoder and 3D-RoFormer++ contributes to enhance the registration robustness and accuracy. }\label{tab:ab_all}
			\vspace{-0.1cm}
			\centering
			\setlength\tabcolsep{2pt}
			\begin{tabular}{l|ccc|ccc|cccc|cccc}
				\toprule
				\multirow{3}*{case} &\multicolumn{6}{c|}{Registration}&\multicolumn{8}{c}{Candidate retrieval}\\
				&\multicolumn{3}{c|}{Mulran}  		&\multicolumn{3}{c|}{KITTI-loop} &\multicolumn{4}{c|}{KITTI}  		&\multicolumn{4}{c}{Campus}\\
				&\multirow{1}*{RR(\%)}  &\multirow{1}*{RRE($^\circ$)}  &\multirow{1}*{RTE(cm)}
				&\multirow{1}*{RR(\%)}  &\multirow{1}*{RRE($^\circ$)}  &\multirow{1}*{RTE~(cm)}
				&\multirow{1}*{AUC} &\multirow{1}*{F1max}	  &\multirow{1}*{Recall@1}  &\multirow{1}*{Recall@1\%} 
				&\multirow{1}*{AUC} &\multirow{1}*{F1max}	  &\multirow{1}*{Recall@1}  &\multirow{1}*{Recall@1\%}  	\\
				
				\midrule	\rowcolor{gray!30}
				full &\textbf{98.22} &\textbf{0.17} &\textbf{7.4} 
				&\textbf{100} &\textbf{0.21} &\textbf{4.3}
				&0.958&0.922 		&0.937 	&\textbf{0.993}		
				&0.972&\textbf{0.920}  	&{0.932}&0.987 \\
				
				VoteEncoder w/o &85.46	 	&0.41 			&20.1  		
				&99.99 &0.24&6.6
				&\textbf{0.975}&\textbf{0.940}  		 &\textbf{0.948} &0.990
				&\textbf{0.973}&{0.917}	 	&{0.927}&0.981\\
				
				3D-RoFormer++ w/o &93.67 &0.22 &10.1 		
				&\textbf{100} 	&0.26  &4.8
				&0.950&0.919 &0.920 &0.962
				&0.951&0.871 		&\textbf{0.939} &\textbf{0.994}\\
				\bottomrule	
			\end{tabular}
		\end{subtable}
		
		\vspace{0.1cm}
		\begin{subtable}{0.99\linewidth}
			\caption{\textbf{Training strategy}. Linear probe ensures the best registration performance while allowing for enhanced candidate retrieval performance. }\label{tab:ab_train}
			\vspace{-0.1cm}
			\centering
			\setlength\tabcolsep{2pt}
			\begin{tabular}{l|ccc|ccc|ccc|cccc|cccc}
				\toprule
				\multirow{3}*{case} &\multirow{3}*{b}&\multirow{3}*{p}&\multirow{3}*{n}&\multicolumn{6}{c|}{Registration}&\multicolumn{8}{c}{Candidate retrieval}\\
				&&&&\multicolumn{3}{c|}{Mulran}  		&\multicolumn{3}{c|}{KITTI-loop} &\multicolumn{4}{c|}{KITTI}  		&\multicolumn{4}{c}{Campus}\\
				&&&&\multirow{1}*{RR(\%)}  &\multirow{1}*{RRE($^\circ$)}  &\multirow{1}*{RTE(cm)}
				&\multirow{1}*{RR(\%)}  &\multirow{1}*{RRE($^\circ$)}  &\multirow{1}*{RTE~(cm)}
				&\multirow{1}*{AUC} &\multirow{1}*{F1max}	  &\multirow{1}*{Recall@1}  &\multirow{1}*{Recall@1\%} 
				&\multirow{1}*{AUC} &\multirow{1}*{F1max}	  &\multirow{1}*{Recall@1}  &\multirow{1}*{Recall@1\%}  	\\
				
				\midrule	\rowcolor{gray!30}
				linear probe&6&6&6 &\textbf{98.22} &\textbf{0.17} &\textbf{7.4} 
				&\textbf{100} &\textbf{0.21} &\textbf{4.3}
				&\textbf{0.958}&\textbf{0.922} 		&\textbf{0.937} 	&\textbf{0.993}		
				&\textbf{0.972}&\textbf{0.920}  	&\textbf{0.932}		&\textbf{0.987} \\
				
				fine-tune &1&1&1&90.13	 	&0.18			&8.1  		
				&\textbf{100} &\textbf{0.21}&4.5
				&0.900&{0.834}  		 &{0.911} &0.975
				&0.883&{0.818}	 	&{0.877}&0.971\\
				
				{end-to-end} &1&1&1
				&85.70 &0.21 &9.6		
				&\textbf{100} 	&0.22  &4.7
				&0.868&0.808 &0.872 &0.966
				&0.834&0.782 		&0.871 &0.979\\
				\bottomrule

			\end{tabular}
		\end{subtable}
		\begin{tablenotes} 
			\item Default structure are marked in \colorbox{gray!30}{gray} and the best results are highlighted in bold.
		\end{tablenotes} 
		
		\vspace{-0.4cm}
\end{table*}}

\newcommand{\VoteTab}{
	\begin{table}
		\caption{A Comprehensive Evaluation of VoteEncoder on Continuous Registration Task Using the KITTI Dataset.}
		\setlength\tabcolsep{1.5pt}
		\renewcommand\arraystretch{1}

		\begin{tabular}{l|ccc|cccccc}
			\toprule
			\multirow{2}*{Model} 
			&\multicolumn{3}{c|}{Direct metrics} &\multicolumn{6}{c}{Indirect metrics}  \\
			&{RR(\%)}  &{RRE($^\circ$)}  &{RTE(cm)} & IR & MR & HR	&{PIR} & PMR & PHR \\
			
			\midrule
			full	
			&\textbf{99.82} 		&\textbf{0.17} 			&\textbf{3.9}  		 &{78.2}&\textbf{9.2}&\textbf{69.2}	&{80.9}	&\textbf{62.8}&\textbf{86.3}\\
			VoteEncoder	w/o
			&\textbf{99.82} 	&{0.22} 	&{6.4}  	&\textbf{84.5}&3.7&35.1	&\textbf{97.0}	&{15.7}&43.2\\
			{5KPEncoder}	&\textbf{99.82} &{0.20} 	&{4.6}  	&{81.2}&6.3&51.0	&{88.6}	&{39.4}&77.5\\
			\bottomrule	
		\end{tabular}
		\begin{tablenotes} 
			\item The best results are highlighted in bold.
		\end{tablenotes} 
		
		\vspace{-0.4cm}
		\label{tab:Vote}
\end{table}}

%% file: figure.tex
\newcommand{\motiFig}{
	\begin{figure}[t] 
		\centering
		\includegraphics[width=0.99\linewidth]{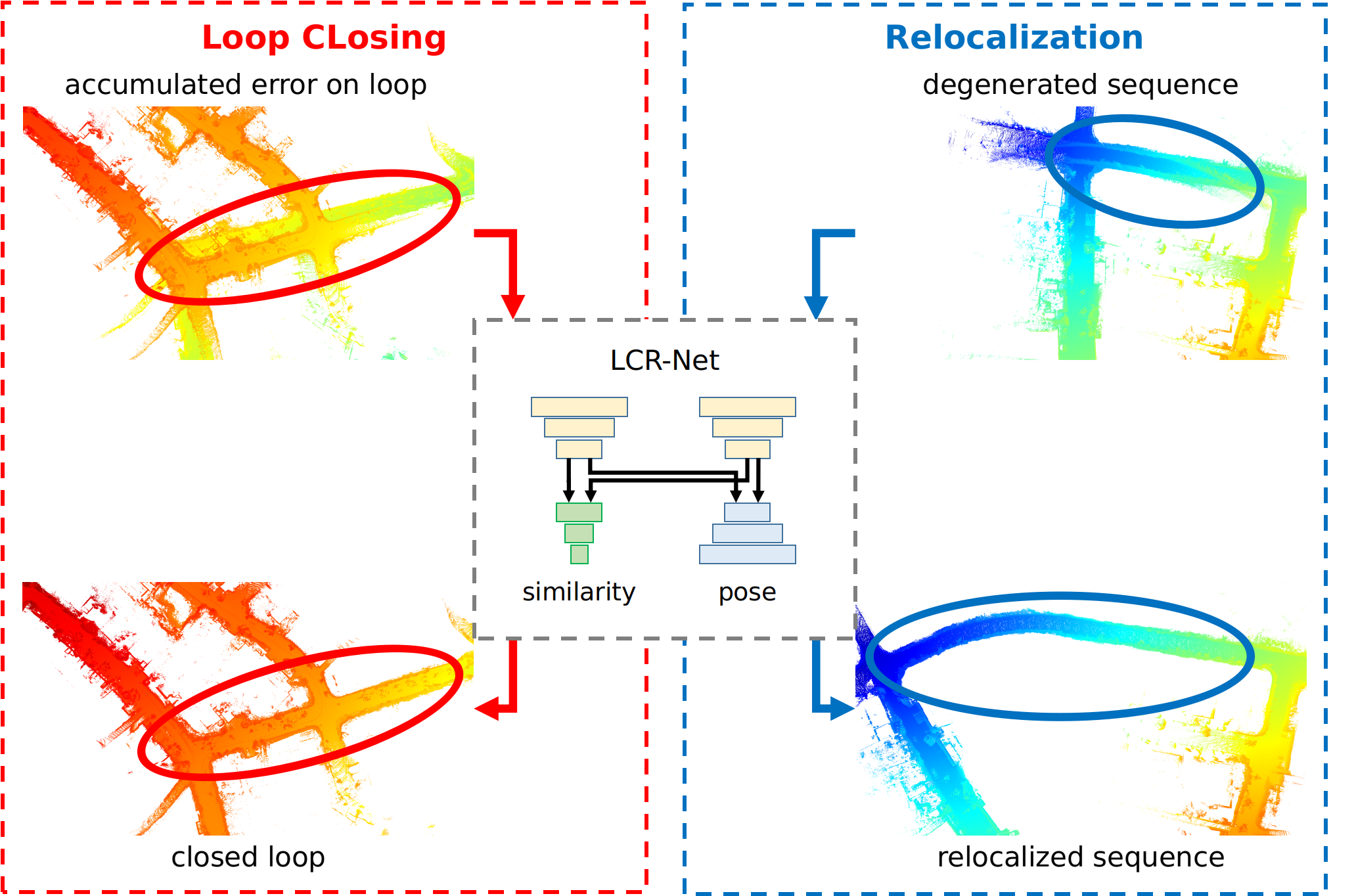}
		\caption{Our proposed \name{} for loop closing and relocalization. \name{} solves both tasks by first retrieving the most similar frame from the map and then estimating the 6-DoF pose.
		}
		\label{fig:motivation} 
		\vspace{-0.4cm}
\end{figure}}

\newcommand{\frameworkFig}{\begin{figure}[t] 
		\centering
		\begin{subfigure}[b]{0.24\textwidth}
			\includegraphics[width=0.95\textwidth]{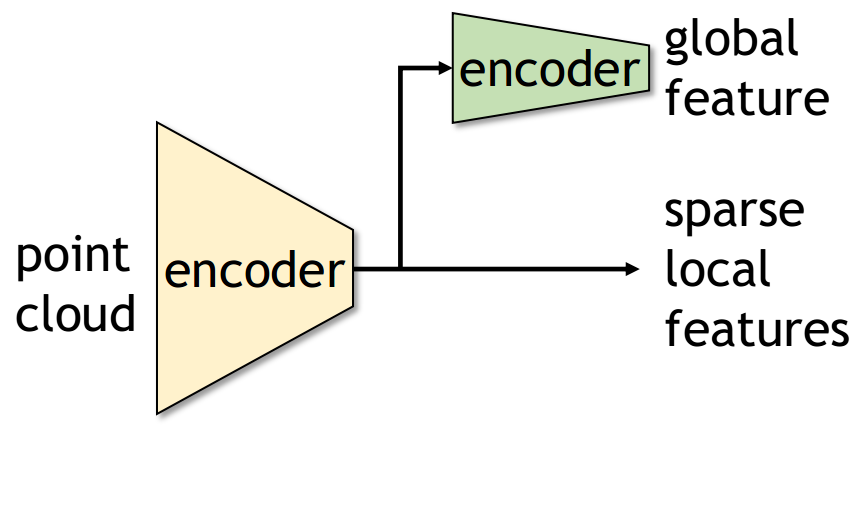}
			\caption{Common framework}
			\label{fig:frame1}
		\end{subfigure}
		\begin{subfigure}[b]{0.24\textwidth}
			\includegraphics[width=0.95\textwidth]{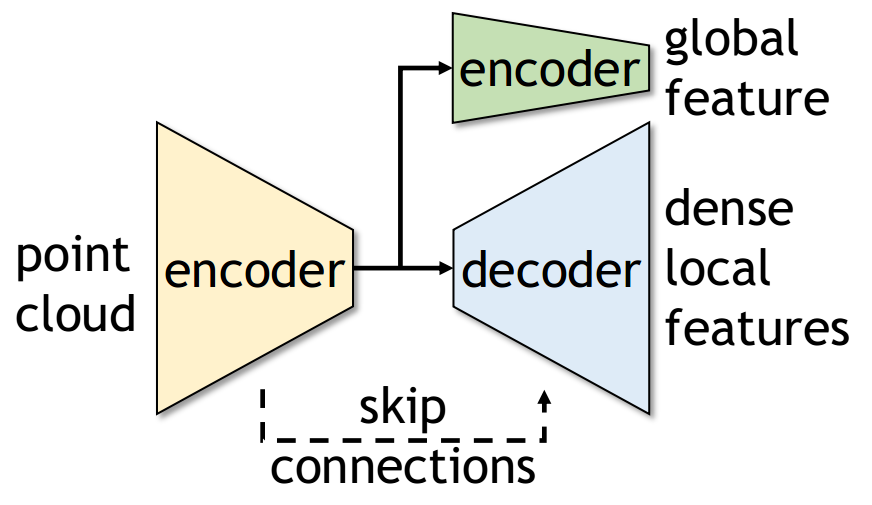}
			\caption{Our framework}
			\label{fig:frame2}
		\end{subfigure}
		
		\caption{Comparison of common framework and our framework. Our framework introduces decoder with skip connections for denser local feature generation.}
		
		\label{fig:framework} 
		\vspace{-0.4cm}
\end{figure}}

\newcommand{\pipelineFig}{
	\begin{figure*}[t]
		\centering
		\begin{overpic}[width=0.90\linewidth]{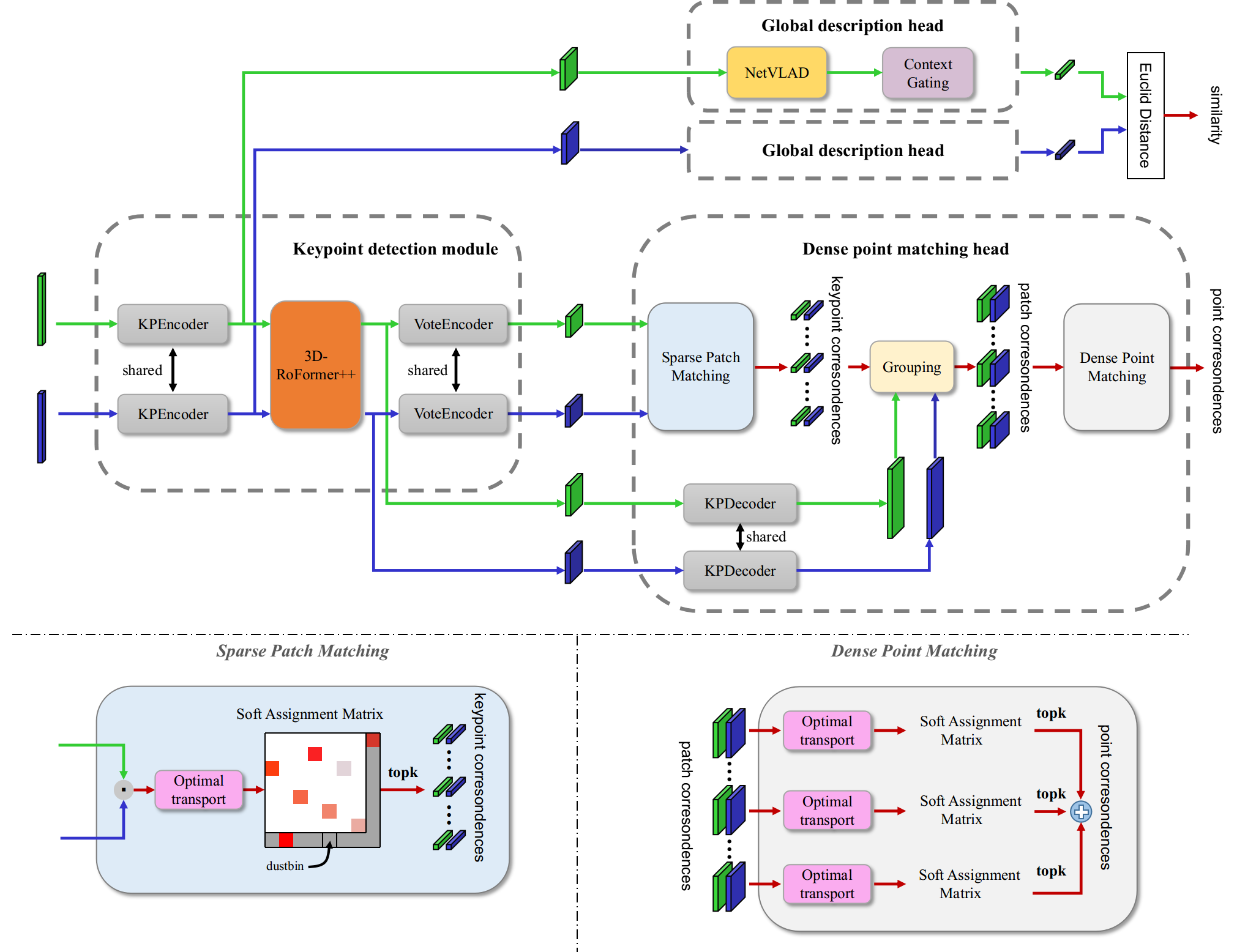}
			\put(0,47.5){{\scriptsize $\mathcal{P}^{\mii A}\!,\! N^{\mii A}\!\!\times\!\! 3$}}
			\put(0,38){{\scriptsize $\mathcal{P}^{\mii B}\!,\! N^{\mii B}\!\!\times\!\! 3$}}
			
			\put(42.5,48){{\scriptsize $\mtf{H}^{\mii A}\!,\! M^{\mii A}\!\!\times\!\! 512$}}
			\put(42.5,41){{\scriptsize $\mtf{H}^{\mii B}\!,\! M^{\mii B}\!\!\times\!\! 512$}}
			
			\put(42,33.5){{\scriptsize $\tilde{\mathbf{F}}^{\mii A}\!,\! \hat{N}^{\mii A}\!\!\times\!\! 256$}}
			\put(42,27.5){{\scriptsize $\tilde{\mathbf{F}}^{\mii B}\!,\! \hat{N}^{\mii B}\!\!\times\!\! 256$}}
			
			\put(42,68){{\scriptsize $\hat{\mathbf{F}}^{\mii A}\!,\! \hat{N}^{\mii A}\!\!\times\!\! 1024$}}
			\put(42,62){{\scriptsize $\hat{\mathbf{F}}^{\mii B}\!,\! \hat{N}^{\mii B}\!\!\times\!\! 1024$}}
			
			\put(82.5,69){{\scriptsize $\m V^{\mii A}\!,\! 1\!\!\times\!\! 256$}}
			\put(82.5,62){{\scriptsize $\m V^{\mii B}\!,\! 1\!\!\times\!\! 256$}}
			
			\put(64.8,34){{\scriptsize ${\mathbf{F}}^{\mii A}\!,\! {N}^{\mii A}\!\!\times\!\! 32$}}
			\put(76.8,34){{\scriptsize ${\mathbf{F}}^{\mii B}\!,\! {N}^{\mii B}\!\!\times\!\! 32$}}
			
			\put(67.1,39.5){{\scriptsize $\mtc{C}$}}
			\put(97.5,40.5){{\scriptsize $\mtc{M}$}}
			
			\put(2,16.5){{\scriptsize $\mtf{H}^{\mii A}$}}
			\put(2,8.5){{\scriptsize $\mtf{H}^{\mii B}$}}
			\put(38.2,5.8){{\scriptsize $\mtc{C}$}}
			\put(31.3,18.7){{\scriptsize $\mathbf{C}$}}
			
			\put(79.8,16.7){{\scriptsize $\mtf{Z}_i$}}
			\put(79.8,10.3){{\scriptsize $\mtf{Z}_i$}}
			\put(79.8,4.35){{\scriptsize $\mtf{Z}_i$}}
			\put(88.5,5){{\scriptsize $\mtc{M}$}}
			
		\end{overpic}
		\caption{\textbf{Pipeline overview.} 
			\name{} consists of three main components: a keypoint detection module, a global description head, and a dense point matching head. The keypoint detection module extracts three types of features for further processing in two heads. The global description head generates a global descriptor for fast candidate retrieval. The dense point matching head exploits a sparse-to-fine approach to establish dense point matching for 6-DoF pose estimation.} 
		\label{fig:model}
		\vspace{-0.4cm}
\end{figure*}}

\newcommand{\voteencoderFig}{
	\begin{figure}[t] 
		\centering
		\begin{overpic}[width=0.99\linewidth]{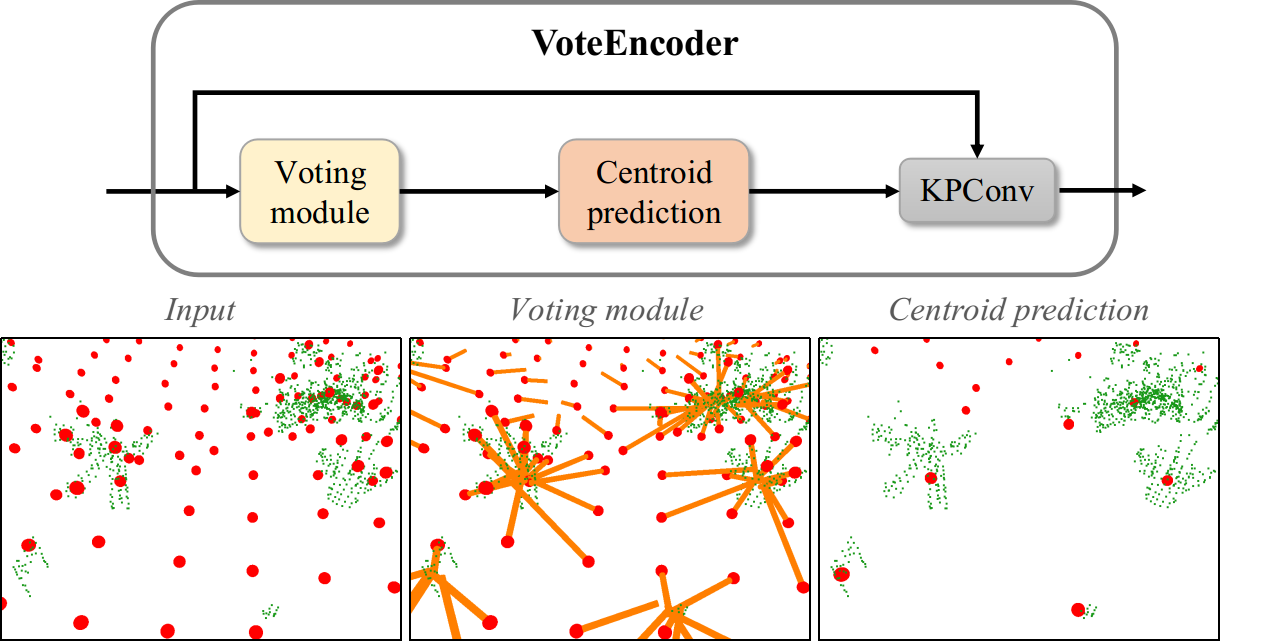}
			\put(0,37.5){{\scriptsize $[\hat{\mathcal{P}},\tilde{\mathbf{F}}]$}}	
			\put(33,40){{\scriptsize ${\mathcal{S}}$}}	
			\put(61,40){{\scriptsize $\hat{\mathcal{S}}$}}	
			\put(91,37.5){{\scriptsize $[\hat{\mathcal{S}},{\mathbf{H}}]$}}
			
			\put(78,45){{\scriptsize $[\hat{\mathcal{P}},\tilde{\mathbf{F}}]$}}		
		\end{overpic}
		\caption{The VoteEncoder takes sparse features as input, and generates offset from each keypoint to its nearest significant region. The final keypoints are estimated using centroid prediction algorithm, and their features are aggregated by KPConv. The ground points are removed from the visualization for clarity.}
		\label{fig:voteencoder} 
		\vspace{-0.4cm}
	\end{figure}
}

\newcommand{\regFig}{\begin{figure*}[p] 
		\centering
		\includegraphics[width=0.99\linewidth]{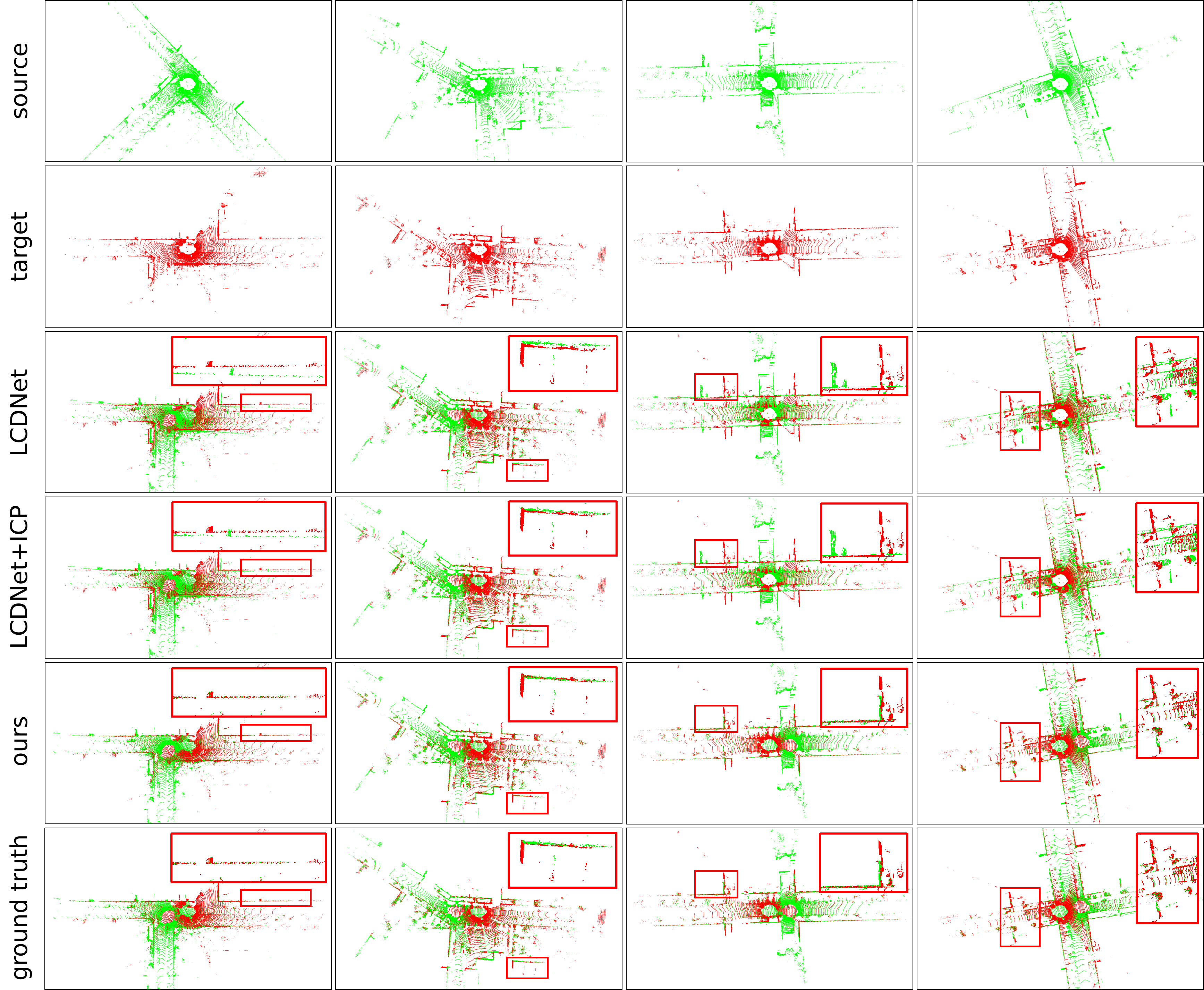}
		\caption{Qualitative comparison of registration using LCDNet and ICP with using LCDNet as initial guess and \name{}. The first two columns depict the successful cases of both methods, where \name{} achieves better alignment. The last two columns show the failure cases of LCDNet, yet \name{} still perfectly aligns the two point clouds.}
		\label{fig:reg} 
		\vspace{4mm}
		\includegraphics[width=0.99\linewidth]{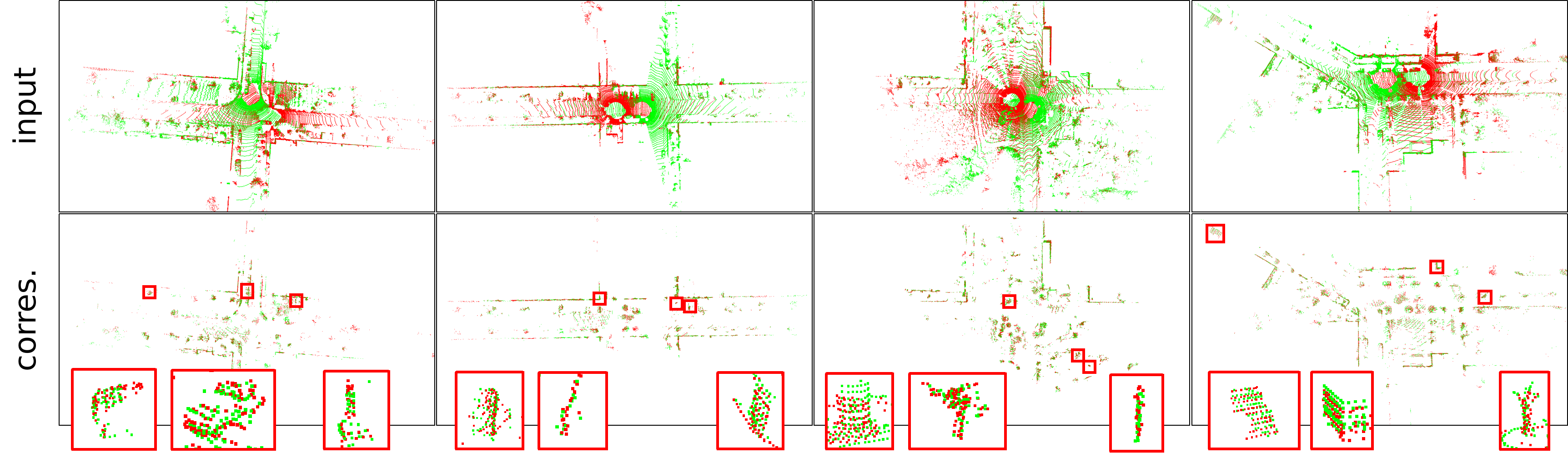}
		\caption{We visualize the correspondent points founded by \name{}. For better visualization, we align the input point clouds and the founded correspondent points using \name{} based on LGR. For each cases, we zoom in on three representative regions for closer examination. Our \name{} effectively finds correspondences covering overall overlapping region and focuses on geometrically significant region.}
		\label{fig:corr} 
\end{figure*}}

\newcommand{\corrcompFig}{\begin{figure}[t] 
		\centering
		\begin{subfigure}[b]{0.45\textwidth}
			\includegraphics[width=\textwidth]{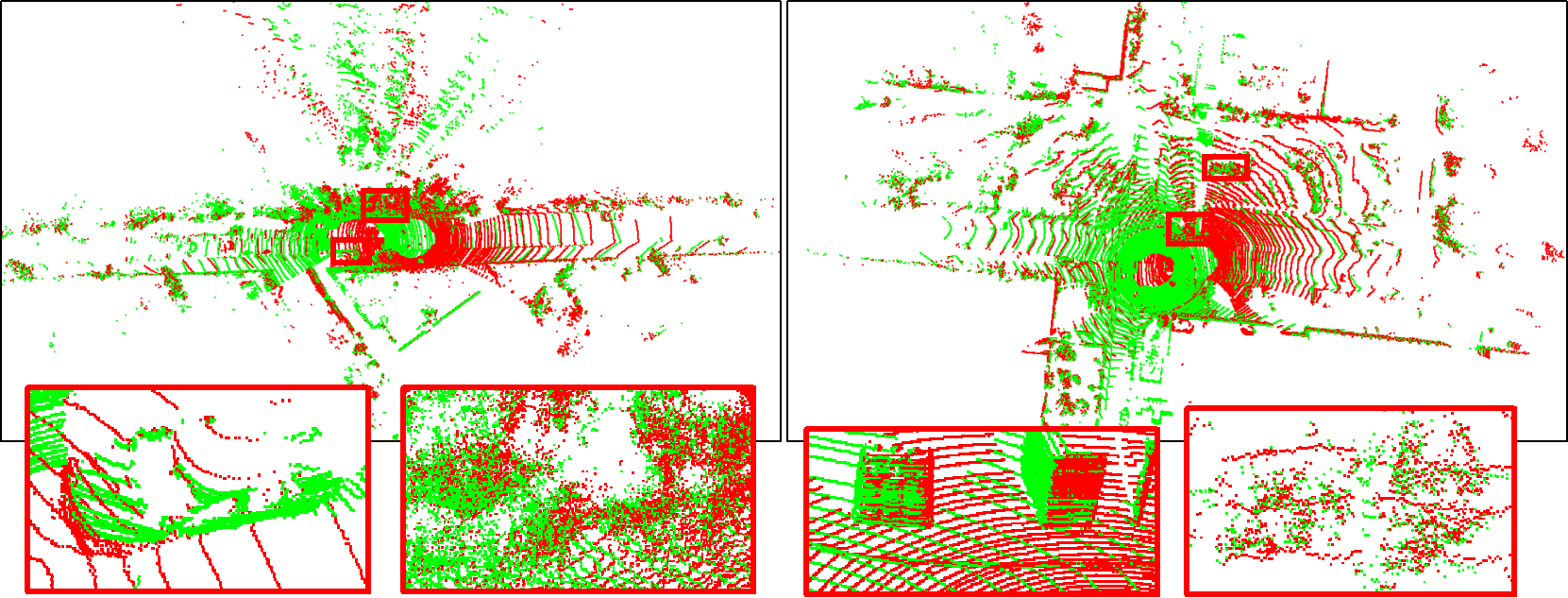}
			\caption{Input point clouds.}
			\label{fig:subfig1}
		\end{subfigure}
		\begin{subfigure}[b]{0.45\textwidth}
			\includegraphics[width=\textwidth]{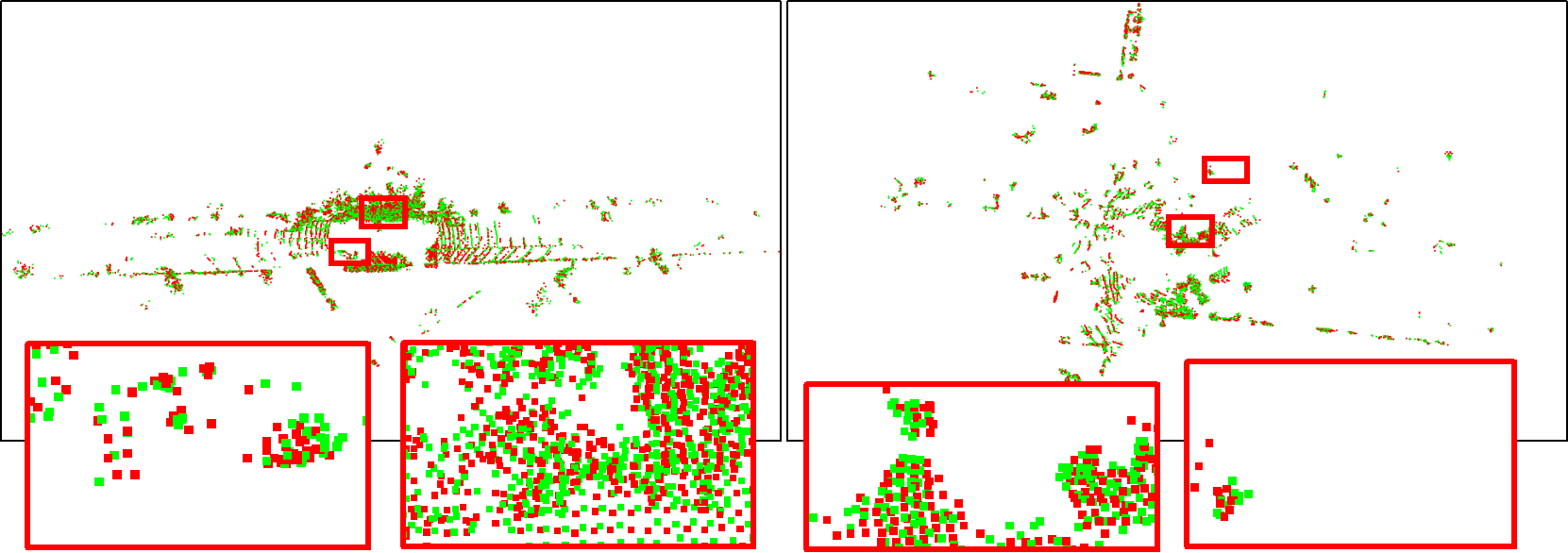}
			\caption{Matching points found by \name{} ablating VoteEncoder.}
			\label{fig:subfig2}
		\end{subfigure}
		\begin{subfigure}[b]{0.45\textwidth}
			\includegraphics[width=\textwidth]{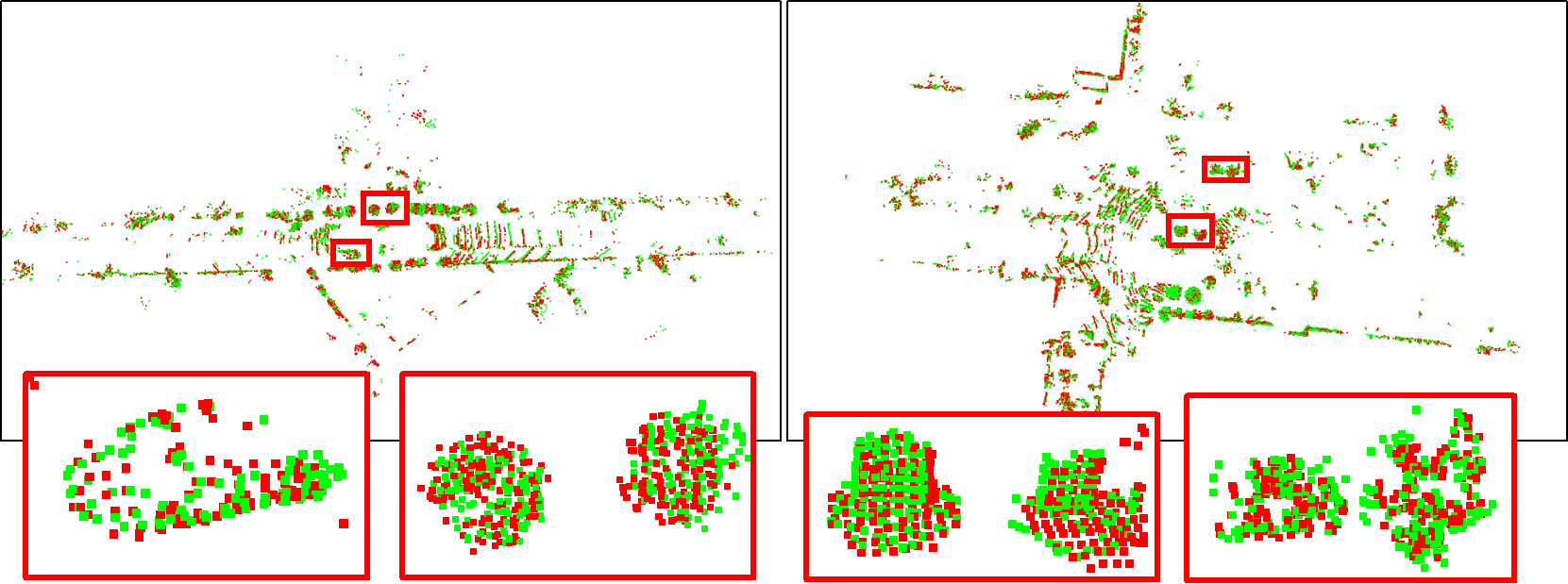}
			\caption{Matching points found by \name{}.}
			\label{fig:subfig3}
		\end{subfigure}
		\caption{We visualize the matching points identified by \name{} to demonstrate the focus of our \name{}. The matching points obtained by \name{} after ablating VoteEncoder is also visualized for comparison. For better visualization, we align the points in two scans using the pose estimated by \name{} based on LGR. In each case, we zoom in on three representative regions for closer examination.}
		\label{fig:corr_comp} 
		\vspace{-0.4cm}
\end{figure}}

\newcommand{\slamFig}{\begin{figure}[t]
		\centering
		\includegraphics[width=0.49\textwidth]{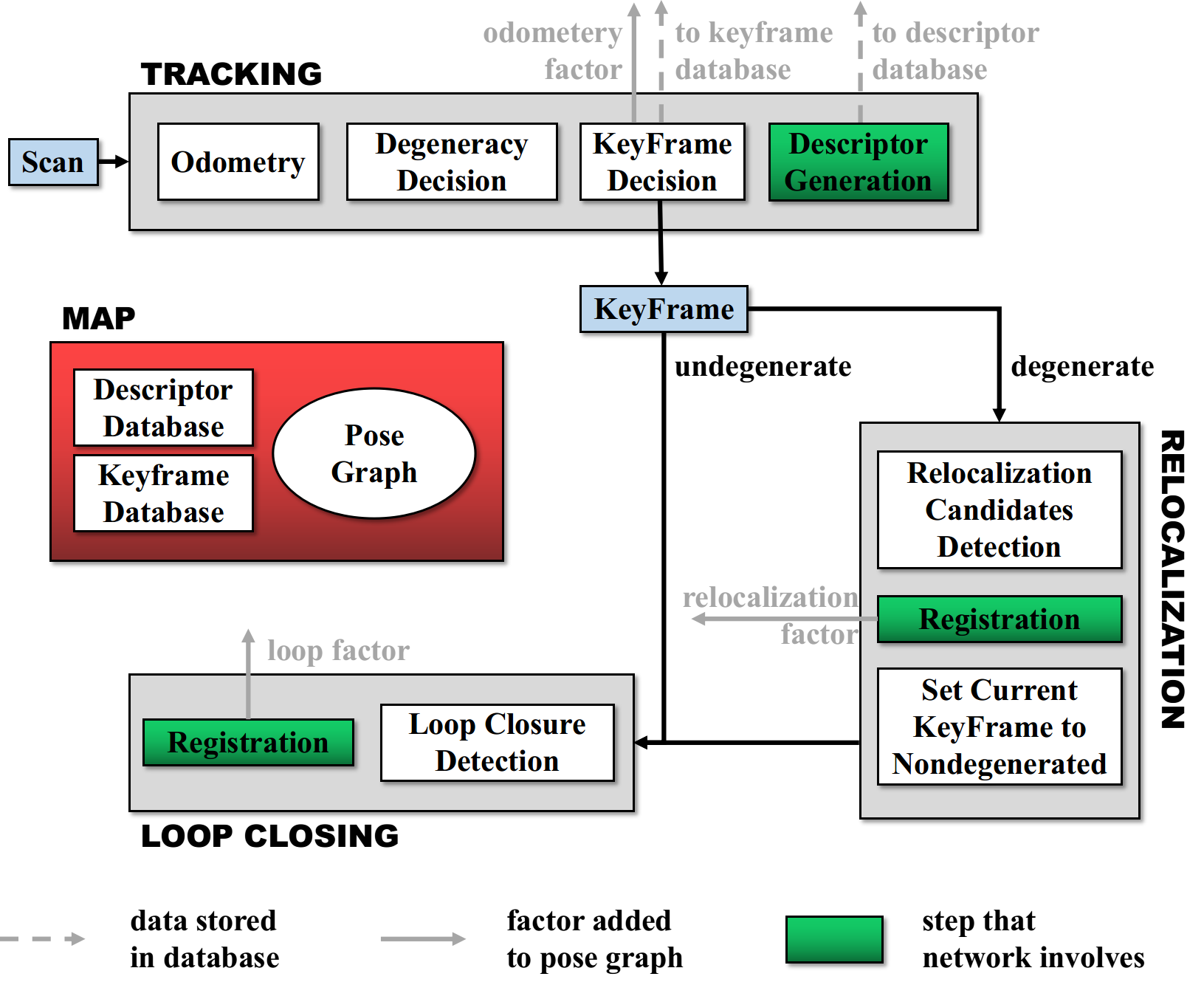}
		\caption{Overview of the laser SLAM system integrated with \name{}, showing key steps executed by the tracking, relocalization and loop closing threads. \name{} serves for relocalization and loop closing.}
		\label{fig:slam} 
		\vspace{-0.4cm}
\end{figure}}

\newcommand{\slamresultFig}{\begin{figure*}[t] 
		\centering
		\begin{subfigure}[b]{0.20\textwidth}
			\centering
			\includegraphics[height=2.5\textwidth]{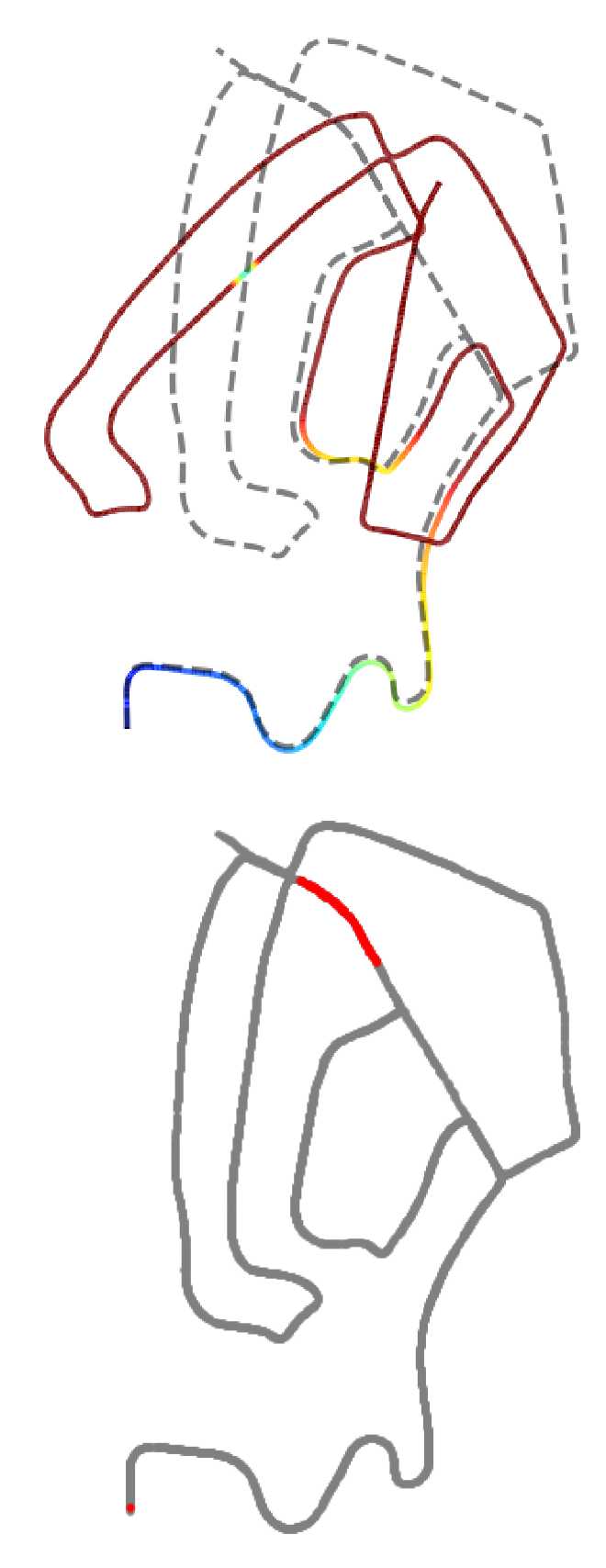}
			\caption{}
			\label{fig:slam1}
		\end{subfigure}
		\hspace{0.3cm}
		\begin{subfigure}[b]{0.20\textwidth}
			\centering
			\includegraphics[height=2.5\textwidth]{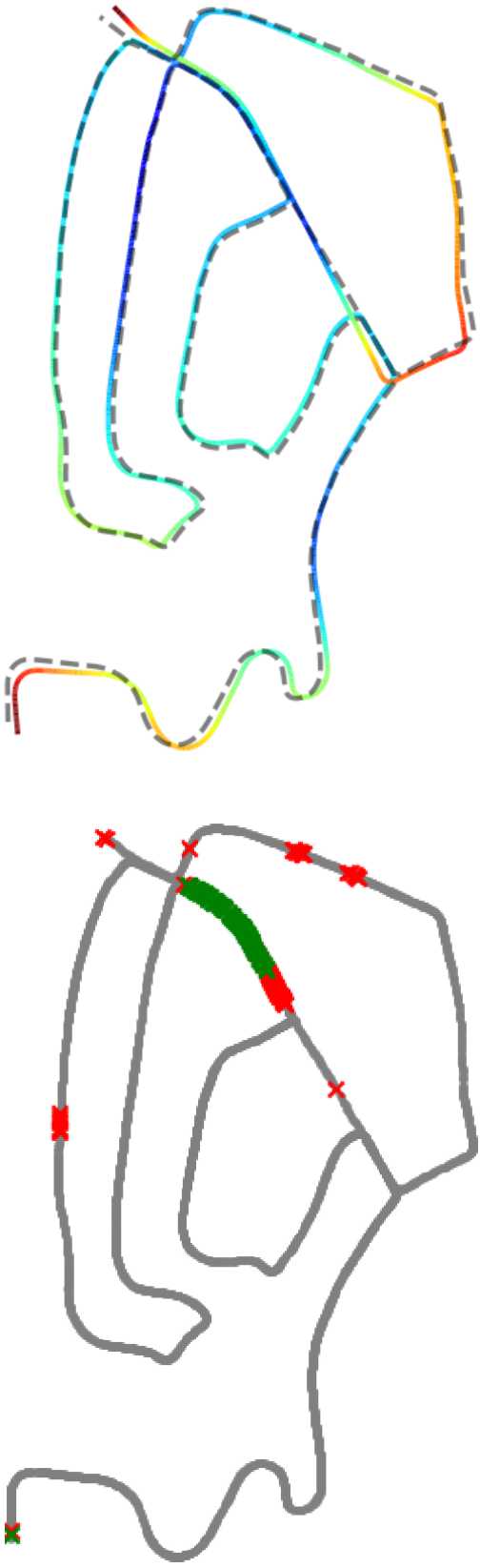}
			\caption{}
			\label{fig:slam2}
		\end{subfigure}
		\hspace{0.3cm}
		\begin{subfigure}[b]{0.20\textwidth}
			\centering
			\includegraphics[height=2.5\textwidth]{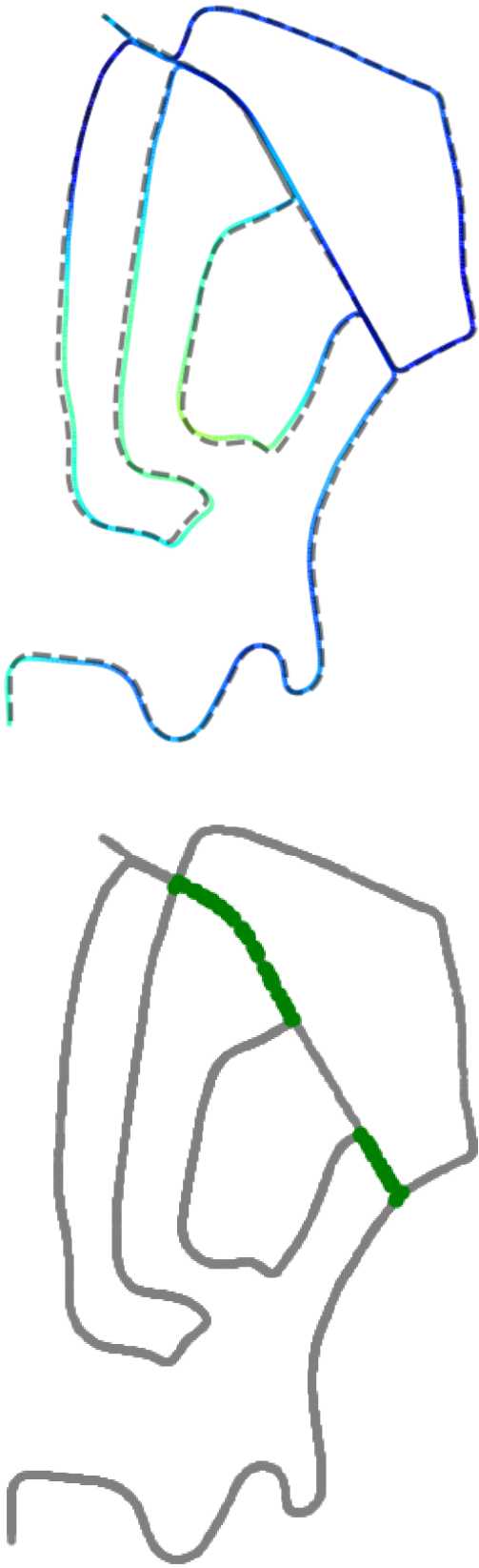}
			\caption{}
			\label{fig:slam3}
		\end{subfigure}
		\hspace{0.3cm}
		\begin{subfigure}[b]{0.20\textwidth}
			\centering
			\includegraphics[height=2.5\textwidth]{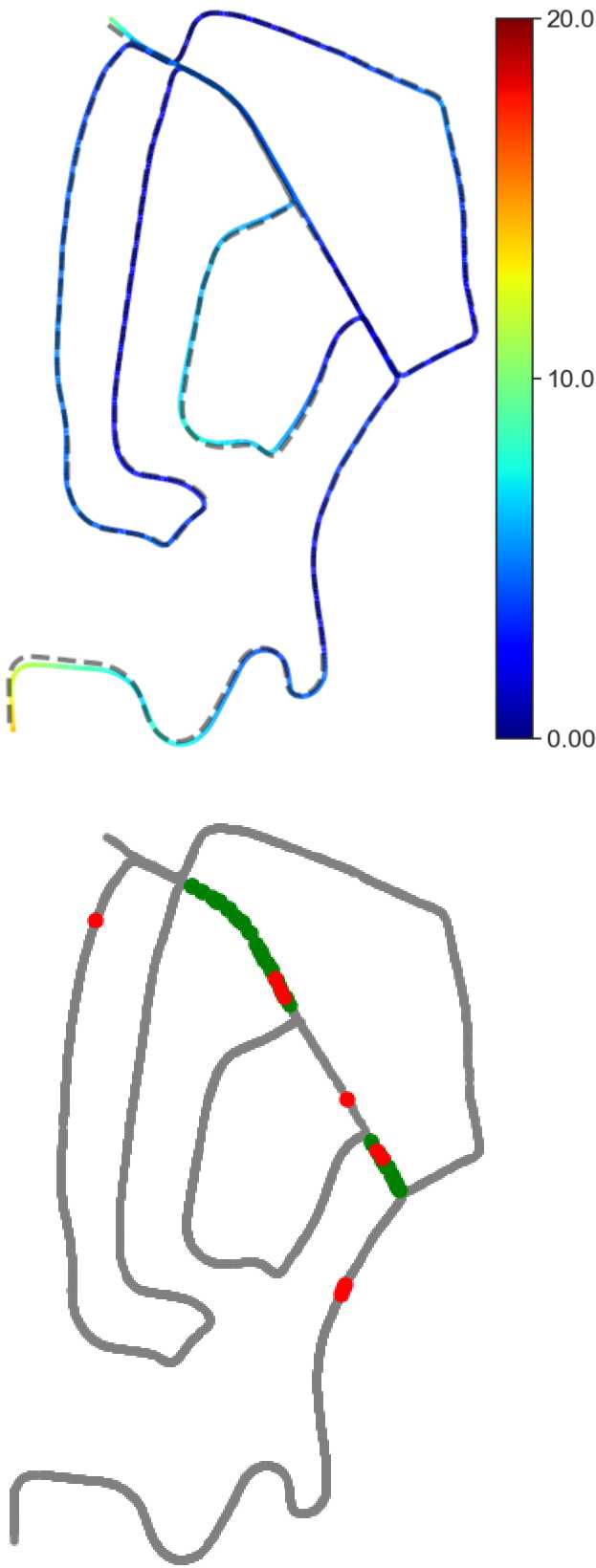}
			\caption{}
			\label{fig:slam4}
		\end{subfigure}
		\caption{Performance of A-LOAM with integrating our approach compared to integrating Scan Context on KITTI sequence 02. We present the results of (a) original A-LOAM (top) and degenerated keyframes colored in red (bottom), (b) A-LOAM with enabling relocalization using our approach (top) and the detected degenerated keyframes (bottom, green crosses \textcolor[rgb]{0,0.5,0}{$\times$} are true positive and red crosses \textcolor{red}{$\times$} are false positive detections), (c) A-LOAM with enabling relocalization and loop closing using our approach (top) and the detected loop closures (bottom, green dots \textcolor[rgb]{0,0.5,0}{$\bullet$} are true positive detections), (d) A-LOAM with enabling relocalization using our approach and loop closing using Scan Context and ICP (top) and the detected loop closures (bottom, green dots \textcolor[rgb]{0,0.5,0}{$\bullet$} are true positive detections and red dots \textcolor[rgb]{1,0,0}{$\bullet$} are false positive).}
		\label{fig:slam_result} 
		\vspace{-0.4cm}
\end{figure*}}

\newcommand{\slamapeFig}{\begin{figure}[t] 
		\centering
		\begin{subfigure}[b]{0.22\textwidth}
			\includegraphics[height=1.3\textwidth]{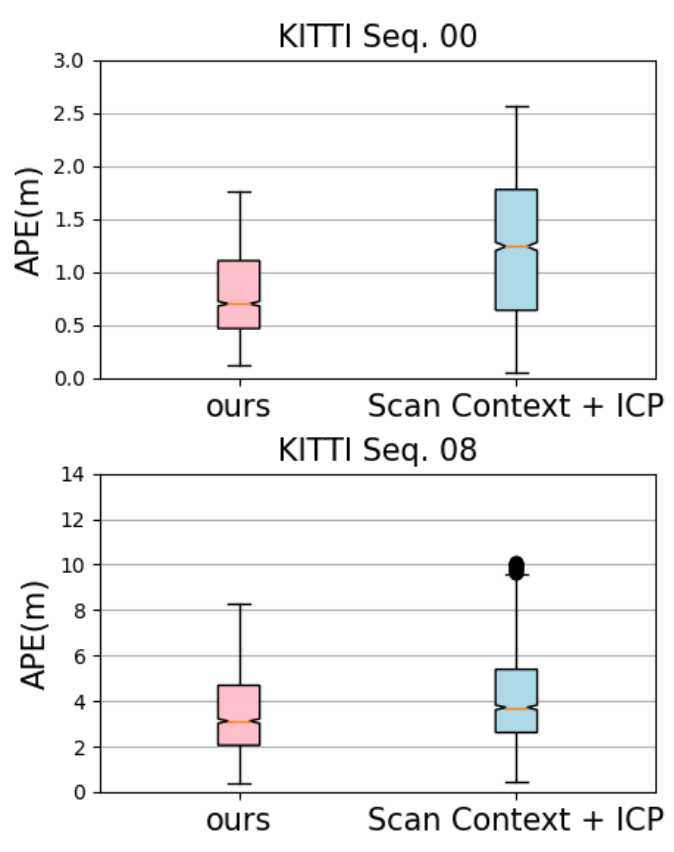}
			\caption{Front end: A-LOAM}
			\label{fig:ape1}
		\end{subfigure}
		\begin{subfigure}[b]{0.22\textwidth}
			\includegraphics[height=1.3\textwidth]{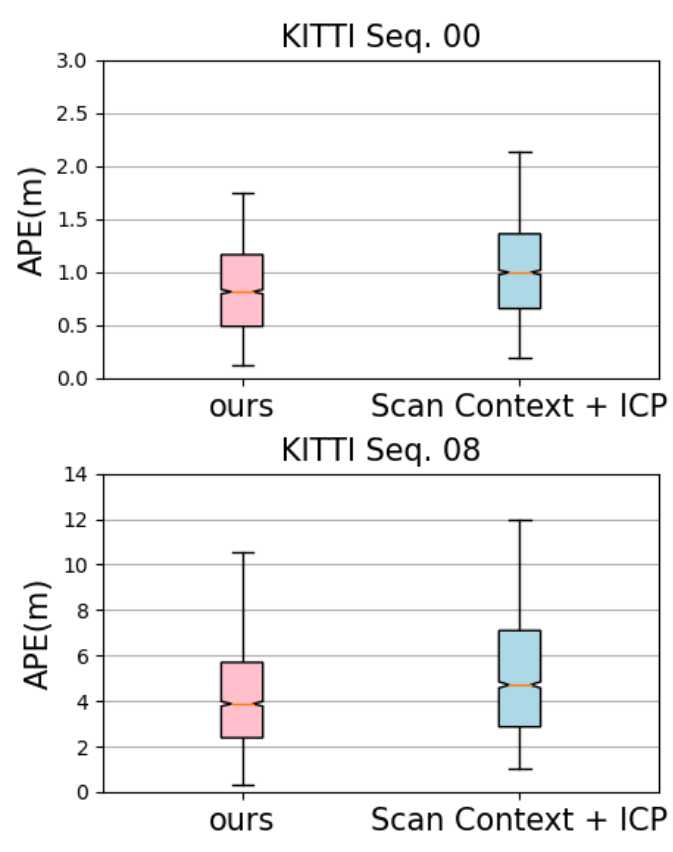}
			\caption{Front end: F-LOAM}
			\label{fig:ape2}
		\end{subfigure}
		\caption{Comparison of SLAM systems with different front-end odometry approaches on different sequences.}
		\label{fig:ape} 
		\vspace{-0.4cm}
\end{figure}}